\documentclass[journal]{IEEEtran}
\ifCLASSINFOpdf
\else
\fi

\usepackage{algpseudocode}
\usepackage{todonotes}
\usepackage{amsmath,amssymb,amsfonts}
\usepackage{graphicx}
\usepackage{subcaption}
\usepackage{textcomp}
\usepackage{xcolor}
\usepackage{setspace} 
\usepackage{multirow}
\usepackage{array}
\newcolumntype{L}{>{$}l<{$}}
\newcolumntype{C}{>{$}c<{$}}
\newcolumntype{R}{>{$}r<{$}}
\newcommand{\Lim}[1]{\raisebox{0.5ex}{\scalebox{0.8}{$\displaystyle \lim_{#1}\;$}}}
\usepackage{booktabs}
\usepackage{algorithm}
\usepackage{colortbl}
\usepackage{url}

\begin{document}
%
\title{Defending Distributed Classifiers Against Data Poisoning Attacks}
%
%
%
\author{Sandamal Weerasinghe, Tansu Alpcan, Sarah M. Erfani and Christopher Leckie
\thanks{S. Weerasinghe and T. Alpcan are with the Department of Electrical and Electronic Engineering, University of Melbourne, Parkville 3010 Victoria, Australia. e-mail: pweerasinghe@student.unimelb.edu.au.}
\thanks{S. M. Erfani and C. Leckie are with the School of Computing and Information Systems, University of Melbourne, Parkville 3010 Victoria, Australia.}
}

%
%

\markboth{Journal of \LaTeX\ Class Files,~Vol.~14, No.~8, August~2015}%
{Shell \MakeLowercase{\textit{et al.}}: Bare Demo of IEEEtran.cls for IEEE Journals}
%



\markboth{}{}
\maketitle

\begin{abstract}
Support Vector Machines (SVMs) are vulnerable to targeted training data manipulations such as poisoning attacks and label flips. By carefully manipulating a subset of training samples, the attacker forces the learner to compute an incorrect decision boundary, thereby cause misclassifications. Considering the increased importance of SVMs in engineering and life-critical applications, we develop a novel defense algorithm that improves resistance against such attacks. Local Intrinsic Dimensionality (LID) is a promising metric that characterizes the outlierness of data samples. In this work, we introduce a new approximation of LID called K-LID that uses kernel distance in the LID calculation, which allows LID to be calculated in high dimensional transformed spaces. We introduce a weighted SVM against such attacks using K-LID as a distinguishing characteristic that de-emphasizes the effect of suspicious data samples on the SVM decision boundary. Each sample is weighted on how likely its K-LID value is from the benign K-LID distribution rather than the attacked K-LID distribution. We then demonstrate how the proposed defense can be applied to a distributed SVM framework through a case study on an SDR-based surveillance system. Experiments with benchmark data sets show that the proposed defense reduces classification error rates substantially (10\% on average).
\end{abstract}


%
\IEEEpeerreviewmaketitle

\section{Introduction}\label{sec:introduction}
\begin{figure*}[ht!]
	\centering
	\begin{subfigure}[t]{0.3\textwidth}
		\includegraphics[width=\textwidth]{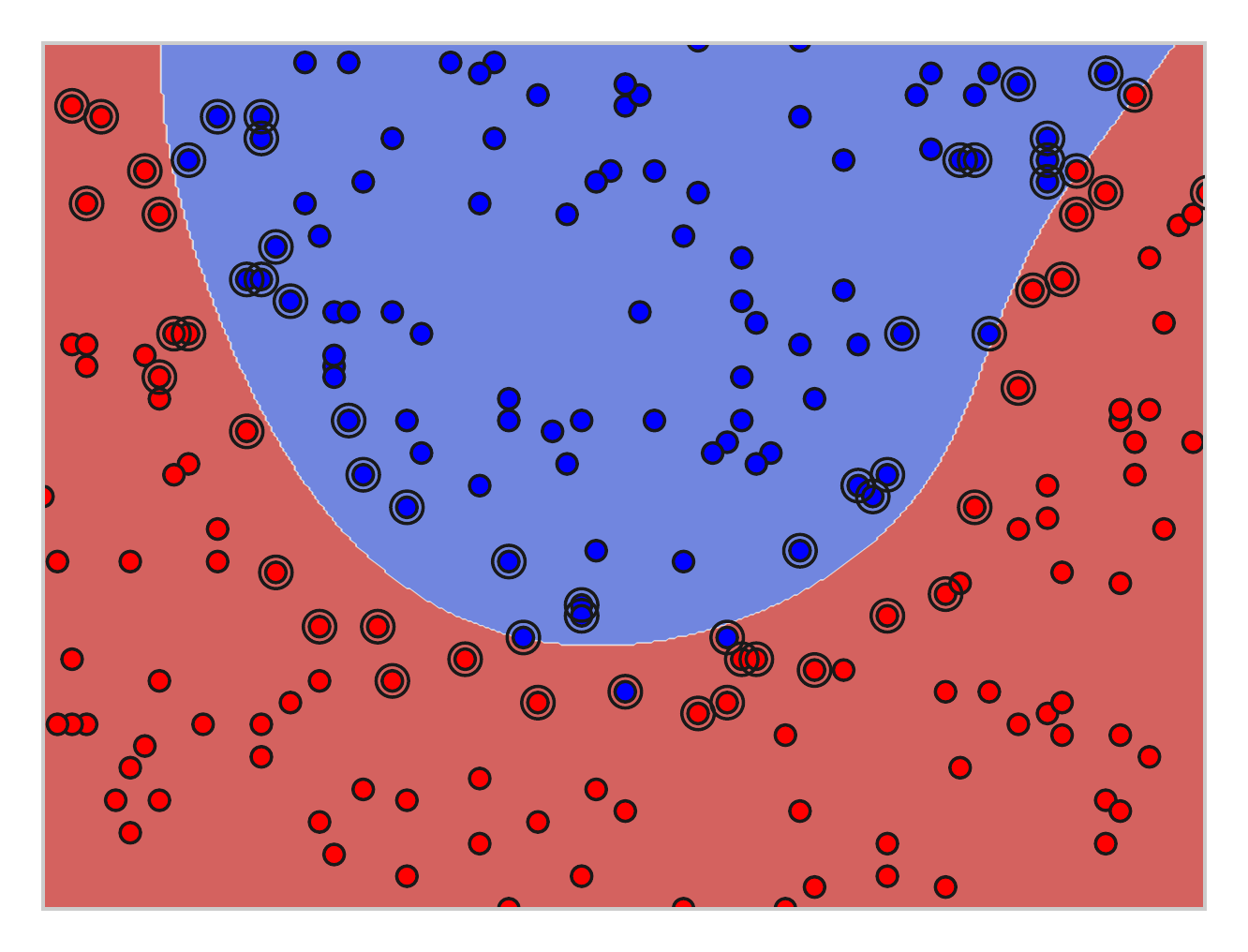}
		\caption{SVM - no attack.}
		\label{fig:toysvm}
	\end{subfigure}
	\begin{subfigure}[t]{0.3\textwidth}
		\includegraphics[width=\textwidth]{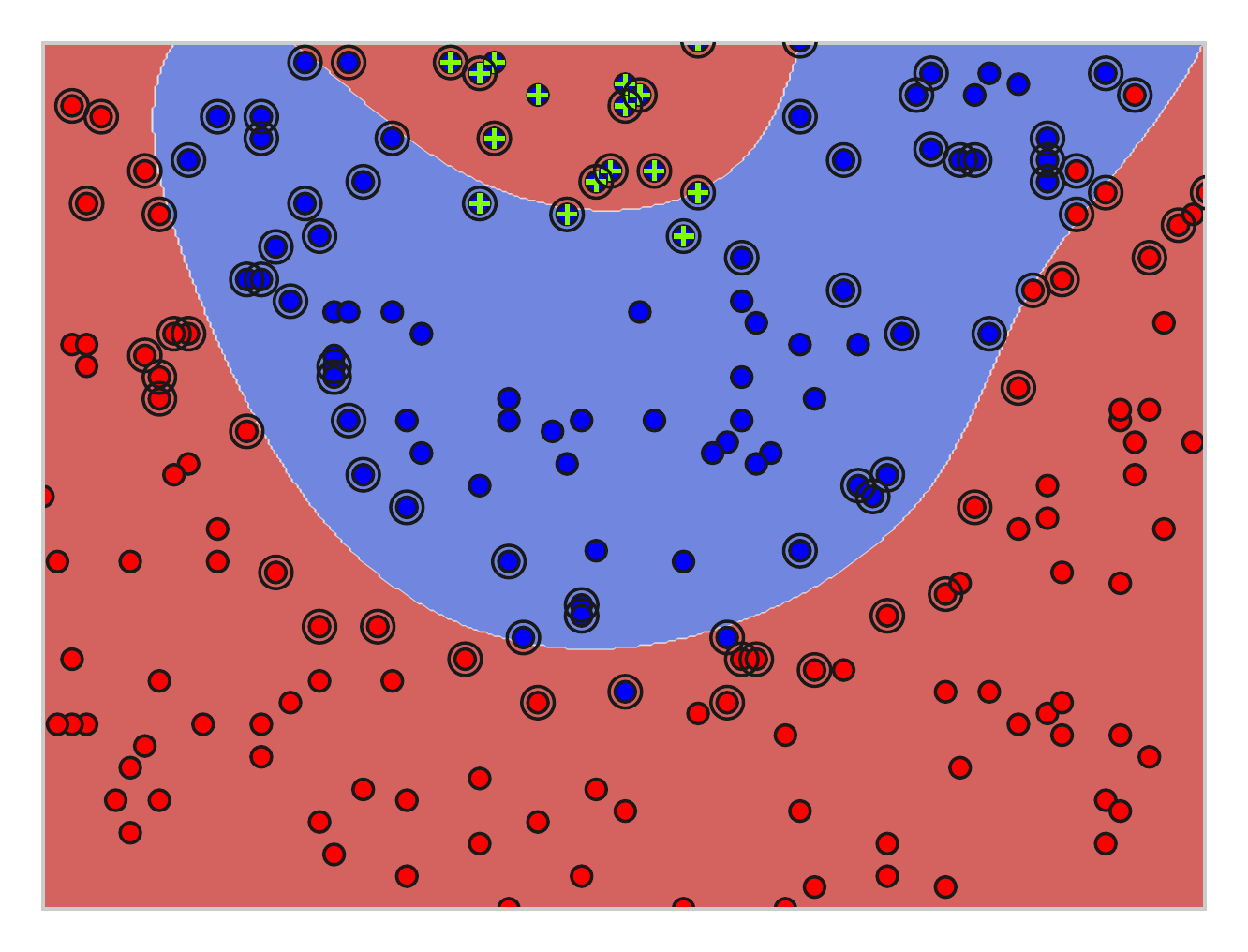}
		\caption{SVM - alfa attack.}
		\label{fig:toyalfa}
	\end{subfigure}
	\begin{subfigure}[t]{0.3\textwidth}
		\includegraphics[width=\textwidth]{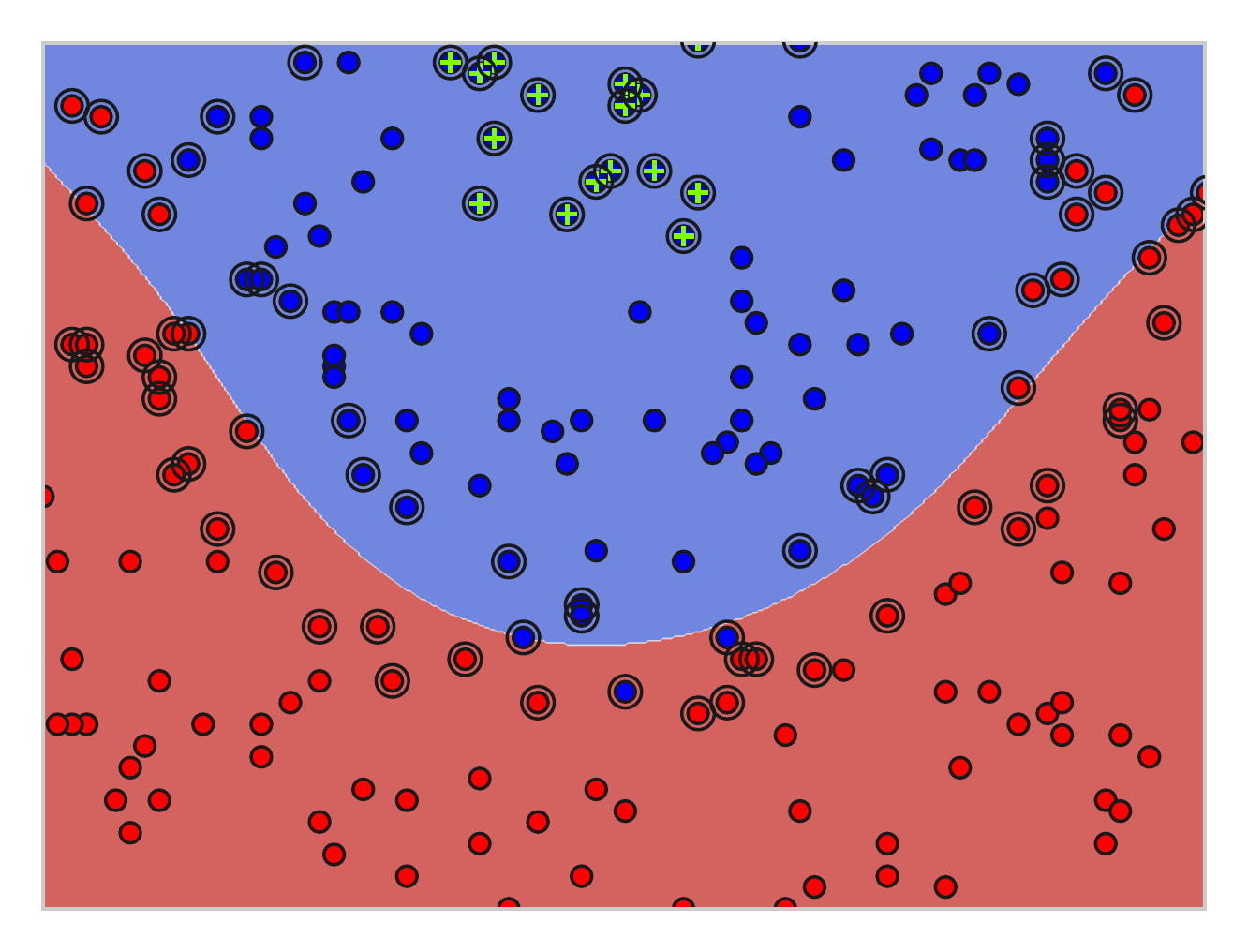}
		\caption{K-LID-SVM - alfa attack.} 
		\label{fig:toylidsvm}
	\end{subfigure}
	\caption{Results on a two dimensional synthetic dataset for an SVM with a RBF kernel ($C=1$ and $\gamma=0.5$). The decision boundary of the SVM trained in the absence of an attack is shown on the left. The middle figure shows the decision boundary of the SVM with no defense when a subset of labels are flipped (from blue to red) using the Adversarial Label Flip Attack (alfa) attack. The figure on the right shows the decision boundary of the K-LID-SVM in the presence of the alfa attack. The flipped samples are shown as green crosses and the support vectors are circled.}
	\label{fig:toy}
\end{figure*}

Recent works in the literature show that even though Support Vector Machines (SVMs) are able to withstand noisy training data by design, maliciously contaminated training data can degrade their classification performance significantly \cite{vorobeychik2018adversarial,dalvi2004adversarial,Biggio:2012:PAA:3042573.3042761, esmaeilpour2019robust}. By carefully manipulating a subset of the training data, the attackers aim to alter the decision boundary of the learner in a way that significantly hinders its prediction capabilities. As Figures, \ref{fig:toysvm} and \ref{fig:toyalfa} show, adversarial manipulations result in a significantly different decision boundary compared to the boundary the classifier would have obtained if the data was pristine. Increasingly, SVMs are being used in engineering and life-critical applications such as autonomous vehicles and smart power grids \cite{sun2002road}. Therefore, it is critical to develop defense mechanisms against training data manipulations (i.e., poisoning or flipping labels) that can subdue their effects.

Prior work in the literature uses data sanitation (i.e., filtering out malicious data) as a defense mechanism when learning in the presence of maliciously altered data \cite{steinhardt2017certified}. Data sanitation may seem like a trivial solution for applications such as image classification under label flipping attacks, where even human experts can detect the flipped samples. But for large data sets under poisoning attacks filtering attacked samples becomes difficult. Expert filtering is also infeasible in other real-world applications such as communication/IoT systems where the data is high dimensional and cannot be transformed into a form that can be easily perceived by humans. Alternatively, embedding the defense into the optimization algorithm is a viable option \cite{suykens1999least,biggio2011support}. However, a majority of such defenses rely on strong assumptions about the data distribution or the attacker.

This paper introduces a novel defense mechanism against poisoning and label flipping attacks using Local Intrinsic Dimensionality (LID), a metric that gives the dimension of the subspace in the local neighborhood of each data sample. Recent evidence suggests a connection between the adversarial vulnerability of learning and the intrinsic dimensionality of the data \cite{amsaleg2017vulnerability, LID_sarah_ICLR}. Previous research has established that LID can be used to identify the regions where adversarial samples lie \cite{LID1_Houle,LID2_Houle}. LID has been applied for detecting adversarial samples in Deep Neural Networks (DNNs) \cite{LID_sarah_ICLR} and as a mechanism to reduce the effect of noisy labels for training DNNs \cite{pmlr-v80-ma18d}.

In this paper, we propose a novel LID estimation called K-LID (that calculates the LID values of data samples in the hypothesis space) that can distinguish attacked samples from benign samples based on the characteristics of the data itself. When data is nonlinearly transformed to a higher dimensional space such that the data from the two classes are well separated, K-LID values of attacked samples would become distinguishable from K-LID values of benign samples. We then use this distinguishing metric to develop a weighted SVM (K-LID-SVM) that is resistant to training data manipulations. Figure \ref{fig:toylidsvm} shows how the proposed defense mechanism can withstand sophisticated attacks with minimal deformations to the decision boundary.

Finally, we assess the applicability of our defense mechanism to a distributed classification setting. The threat of adversarial attacks is aggravated in distributed classification settings as attackers have multiple potential entry points and even if one node is compromised, the effects of the attack can propagate through the entire detection network \cite{zhang2017}. We then evaluate the practical implications of the developed defense scheme through an engineering case-study on a Software-defined radio (SDR) based surveillance system.

Our main \textbf{contributions} are summarized as follows:
\begin{enumerate} 
\item We introduce a novel defense strategy to increase the attack resistance of (distributed) SVMs against poisoning attacks as well as label flipping attacks.
\item The proposed defense uses a LID-based sample weighting mechanism that:\\
- introduces a novel approximation of LID using kernel distances called \textit{K-LID} that calculates the LID values of data samples in the hypothesis space (i.e., a high dimensional transformed space);\\
- uses the likelihood ratio of each data sample (i.e., how many times more likely their K-LID values are from the benign K-LID distribution than the attacked K-LID distribution) as a distinguishing factor and incorporate this knowledge into the SVM training process as weights.
\item We show through numerical experiments conducted with real-world data sets that our proposed approach can increase the attack resistance of SVMs against training time attacks by up to $10\%$ on average.
\item We demonstrate the adaptability of the proposed defense to distributed settings through a case study.
\item We show experimentally that a distributed SVM detection system has 44\% less communication overhead compared to a centralized SVM with only a 3.07\% reduction in detection accuracy on average.
\end{enumerate}

The remainder of the paper is organized as follows. Section \ref{sec:lit_review} provides details of previous literature relevant to this work. Section \ref{sec:attack_model} formally defines the problem being addressed followed by the defense methodology in Section \ref{sec:defense_model}. Section \ref{sec:results} describes the empirical analysis on several real-world datasets followed by the results and discussion. The concluding remarks of Section \ref{sec:conclusions} conclude the paper.

\section{Related Work}\label{sec:lit_review}
In this section, we briefly review previous research findings related to distributed SVMs, training time attacks and defense mechanisms for SVMs.

\subsection{Distributed SVMs}
Distributed training of SVMs has been studied extensively with several variants including cascade SVM, incremental SVM, distributed parallel SVM and consensus-based SVM used in practice \cite{wang2012distributed}. The objective of these variants is to obtain a decision function that is close to the decision function obtained by a centralized SVM with access to all the data points. The approach used by Alpcan and Bauckhage \cite{alpcan2009distributed} decomposes the centralized SVM optimization problem into a set of convex subproblems (one per node) with the nodes having to exchange their respective support vectors (SVs) among each other through a fusion center at each iteration of training. A similar decomposed approach is proposed by Forero et al. \cite{forero2010consensus} for linear SVMs, but instead of exchanging SVs, consensus constraints are added to the classifier parameters.

The latter consensus-based DSVM is used in a series of works by Zhang and Zhu \cite{zhang2017} that aim to design defense strategies for DSVMs against adversaries. The authors use a game-theoretic framework to capture the conflicting interests between the consensus-based DSVM learner and the attacker. Their proposed rejection-based defense strategy prevents updates to the learning models if the primal and dual variables change significantly after training with new data. Their proposed defense model assumes that the initial data it is given is clean and as the primal variables are used in the rejection criteria, this model cannot be used in conjunction with RBF kernels. 

\subsection{SVMs in the Presence of Label Flipping Attacks}
Most works in the literature are concerned with stochastic label noise, rather than malicious attacks. The problem of classification in the presence of label noise has inspired a wide range of research from the machine learning community, see the work of Fr{\'e}nay and Verleysen \cite{frenay2014classification} for a survey. Biggio et al. \cite{biggio2014security} provide a security evaluation of SVMs under both types of training time attacks. Several prior works have shown that SVMs are significantly impacted by mislabeled instances and removing such samples reduces the complexity of SVMs \cite{nettleton2010study}. In the work of Libralon et al. \cite{libralon2009pre}, the authors show that removing flipped samples from training data results in SVMs with simpler decision boundaries.

Zhang and Yang \cite{zhang2003robustness} show that the performance of linear SVMs degrades significantly at merely $5\%$ of flipped labels. Risk minimization under a particular loss function is said to be \textit{label noise-robust} if the probability of misclassification of inferred models with label noise is identical to the same probability without label noise. Manwani and Sastry \cite{manwani2013noise} show that hinge loss is not robust to uniform label noise, consequently, it leads to the conclusion that SVMs are not robust to uniform label noise.

In order to embed a mechanism to consider possible label alterations into SVMs, An and Liang \cite{an2013fuzzy} use fuzzy memberships for each training data point to force data points to influence differently when calculating the separating hyperplane. In the work of Natarajan et al. \cite{natarajan2013learning}, the authors suggest two approaches to modify loss functions to withstand random label noise during training. Liu and Tao \cite{liu2016classification} use traditional loss functions for classification tasks in the presence of random label noise by using importance reweighting.

Suykens and Vandewalle \cite{suykens1999least} introduce Least-Squares SVM (LS-SVM) where a quadratic loss function is used instead of the hinge loss, which results in a non-sparse solution to the optimization problem (all the training samples are assigned non-zero $\alpha$ values). The authors claim this approach prevents the SVM from over-relying on the contribution of certain samples. Label Noise Robust SVM (LN-SVM), proposed by Biggio et al. \cite{biggio2011support}, assumes that the label of each training sample can be independently flipped with the same probability of $\mu$. The probability of label flips is then incorporated into the kernel matrix. With this approach, each training sample is more likely to become a support vector.

\subsection{SVMs under Poisoning Attacks}
Dalvi et al. \cite{dalvi2004adversarial} modeled classification as a game between the classifier and the adversary. They extend the naive Bayes classifier to optimally detect and reclassify perturbed data points, by taking into account the adversary's optimal feature-changing strategy. Zhou et al. \cite{Zhou:2012:ASV:2339530.2339697} introduced an Adversarial SVM (AD-SVM) model which incorporated additional constraint conditions to the binary SVM optimization problem to thwart an adversary's poisoning attacks. Their model only supports data that is linearly separable and leads to unsatisfactory results when the severity of actual attacks differs from the expected attack severity by the model. 

One approach for learning in the presence of poisoned training data is to identify and remove such samples prior to training. Steinhardt et al. \cite{steinhardt2017certified} introduced a framework that uses an outlier detector before training in order to filter out poisoned data. They consider two scenarios, (i) where there is a clean outlier detector (trained independently without being affected by the poisoned data), and (ii) where the outlier detector is also compromised. While the framework performs well in the first scenario, the authors claim that the attacker can subvert the outlier removal and obtain stronger attacks in the second scenario. Laishram and Phoha \cite{laishram2016curie} introduced an algorithm that clusters the data in the input space and utilizes the distances among data points in the same cluster in the (input feature + label) space to identify the outliers. These works can be considered as a pre-processing step and could be used in conjunction with our proposed defense mechanism to further increase the attack resistance of SVMs.

In this work, we focus on both types of training time attacks (i.e., poisoning and label flipping) against SVMs and introduce one defense algorithm that can withstand both types of attacks. To the best of our knowledge, no existing work has explored the use of LID for defense against training time attacks on SVMs. In addition, the impact of adversarial data manipulations on distributed learners is understudied.


\section{Problem Definition}\label{sec:attack_model}
We consider an adversarial learning problem in the presence of a malicious adversary. The adversary's goal is to corrupt the classification model generated in the training phase to maximize the classification error of the detection system. This type of attack is referred to as a \textit{poisoning availability attack} in the adversarial learning literature where the adversary's goal is to affect the prediction results indiscriminately, without seeking specific mispredictions. Take $S:=(X,y)$ to be the labeled training data where $X\in\mathbb{R}^{n\times d}$ and $y_{i}\in\{\pm1\}$ for $i\in\{1,\dots,n\}$. To achieve its goal, the adversary either flips a fraction of training labels ($\tilde{S}:=(X,\tilde{y})$) or perturbs the features of a fraction of data samples ($\tilde{S}:=(\tilde{X},y)$). When the learner trains on the contaminated data, it obtains a distorted decision boundary that is significantly different from the decision boundary it would have obtained if the data was pristine. 

In the following section, we introduce the different attack strategies an adversary may use against an SVM. In Section \ref{sec:defense_model}, we introduce our novel LID based defense models for SVMs against such training time attacks. Refer to Section \ref{sec:results} for details of the experimental setup (i.e., simulation, feature selection and datasets) and empirical evidence proving the effectiveness of the proposed defense.

\subsection{Attack Models}
\subsubsection{Label Flipping Attacks}\label{sec:label_attack_models}
Through a series of works, Xiao et al. \cite{alfa,biggio2011support} introduced several attack models that carefully select a subset of training labels to be flipped in order to maximize an SVM's classification error. The attack models assume that the attacker has perfect knowledge of the attacked system (white-box attack). This means that the attacker is aware of the SVM parameters $C$ and $\gamma$ and training data $S:=(X,y)$. While the attacker's ability is over-estimated, it can be considered as a worst-case scenario for the defender. Moreover, relying on secrecy for security is considered as a poor practice when designing attack resilient learners \cite{biggio2014security}. 

The attack forces the learner to erroneously shift the decision boundary such that there is a significant deviation from an SVM trained on a non-flipped dataset. The attacker is restricted such that it is only allowed to flip the labels of the training data and only a maximum of $L$ label flips are allowed. $L$ bounds the attacker's capability and is fixed \textit{a priori}. The main attack strategies we test against are:
\begin{itemize}
	\item \textbf{Adversarial Label Flip Attack (alfa):}
	take $\tilde{y}$ to be the contaminated training labels. The attack model can be considered as a search problem for $\tilde{y}$ that achieves the maximum difference between the empirical risk for classifiers trained on $\tilde{y}$ and $y$.
	\item \textbf{ALFA based on Hyperplane Tilting (alfa-tilt):} 
	in this attack, the tilt in the separating margin before and after distorting the dataset is used as a surrogate to select the optimal label flips instead of classification error in alfa.
	\item \textbf{Farfirst:}
	samples that are furthest from the separating margin of the non-flipped SVM are flipped.
	\item \textbf{Nearest:}
	samples that are nearest to the separating margin of the non-flipped SVM are flipped.
	\item \textbf{Random:}
	a subset of samples are randomly selected from the training data and flipped.
\end{itemize}

\subsubsection{Poisoning Attacks}\label{sec:attack_model_poisoning}
For binary SVMs, Biggio et al. \cite{Biggio:2012:PAA:3042573.3042761} introduced the poisoning attack algorithm \textbf{(PA)} that perturbs data points in feature space such that there is a maximal increase in the classification error. The authors assume that the attacker has perfect information, resulting in a worst-case analysis for the defenders. The authors address the problem of finding the best attack points by formulating an optimization problem that maximizes the learner's validation error. They empirically demonstrate that the gradient ascent algorithm can identify local maxima of the non-convex validation error function. The main highlight of this work is that it can identify attack points in transformed spaces (using kernel methods).

In contrast, the restrained attack \textbf{(RA)} introduced by Zhou et al. \cite{Zhou:2012:ASV:2339530.2339697} conducts attacks in input space. For each malicious data point $x_{i}$ the adversary aims to alter, it picks a innocuous target $x_{i}^{t}$ and perturbs $x_{i}$ towards $x_{i}^{t}$. The amount of movement is controlled by two parameters, the discount factor ($C_{\xi}$) and the attack severity ($f_{\text{attack}}$).

Li et al.\cite{li2016general} introduced the Coordinate Greedy \textbf{(CG)} attack which is modeled as an optimization problem. The objective of the attacker is to make the perturbed data points appear as benign as possible to the classifier while minimizing the modification cost. For each malicious data point $x_{i}$, the attacker iteratively chooses a feature and greedily updates it to incrementally improve the attacker's utility.

\section{Defense Model for SVMs}\label{sec:defense_model}
We now present the inner workings of our novel LID based defense algorithm for SVMs. The defense algorithm we propose consists of several components. First, the LID values of all the data samples are calculated using a novel LID approximation (K-LID). Then, for each data sample, we calculate how many times more likely its K-LID value is from the K-LID distribution of benign samples than the K-LID distribution of attacked samples, i.e., likelihood ratio (LR). Subsequently, we fit a smooth function to the LR values to be able to predict the LRs for unseen K-LID values. Finally, during SVM training, we weight each sample by the LR function value corresponding to its K-LID value.

First, in Section \ref{sec:background}, we present how SVMs can be distributed among multiple compute nodes by building upon an existing DSVM framework \cite{alpcan2009distributed}. Then, we briefly introduce the theory of LID for assessing the dimensionality of data subspaces. Subsequently, in Section \ref{sec:lid_calculation} we describe in detail each of the novel components of our defense algorithm.

\subsection{Background}\label{sec:background}

\subsubsection{Distributed Weighted SVMs}\label{sec:distributed_lid_svm}
In SVMs, in order to penalize large slack values (i.e., $\xi_i$), a regularization constant $C$ was introduced to the optimization problem. $C$ penalizes all training samples equally. In order to selectively penalize samples, we introduce a weight $\beta_{i}$ for each sample \cite{yang2007weighted}. A small $\beta_{i}$ value would allow for a large $\xi_i$ value and the effect of the sample would be de-emphasized. Conversely, a large $\beta_{i}$ value would force $\xi_i$ to be smaller and therefore the effect of the particular sample would be emphasized. The weighted SVM learning can be formulated as the following convex quadratic programming problem:
\begin{equation}\label{eq:svm_primal}
\begin{aligned}
& \underset{w,\xi_i,b}{\text{minimize}}
& & \frac{1}{2}\|w\|^2+C\sum_{i=1}^n\beta_{i} \xi_i\\
& \text{subject to}
& & y_{i}(\langle\ w,\varPhi(x_i)+b)\rangle \geq 1-\xi_i, &\; i = 1, \ldots, n \\
&   &   & \xi_i \geq 0. \; & i = 1, \ldots, n 
\end{aligned}
\end{equation}
Then the dual formulation of the problem take the form:
\begin{equation}\label{eq:svm_dual}
\begin{aligned}
& \underset{\alpha}{\text{maximize}}
& & \sum_{i=1}^{n}\alpha_{i} - \frac{1}{2}\sum_{i,j=1}^{n}\alpha_{i}\alpha_{j} y_{i} y_{j}k(x_{i},x_{j}) \\
& \text{subject to}
&   & 0\leq\alpha_{i}\leq C \beta_{i}, i = 1, \ldots, n   \\
&   &   & \sum_{i=1}^{n}\alpha_{i} y_{i}=0.
\end{aligned}
\end{equation}

In order to decompose the centralized SVM classification problem into sub problems, define a set of $\mathcal{M}:=\{1,2,\ldots,M\}$ distributed compute units with access to different subsets, $S_{i}$, $i \in \mathcal{M}$, of the labeled training data such that $S = \bigcup_{i=1}^{M} S_{i}$. Given this partition, define the vectors $\{\alpha^{(1)},\alpha^{(2)},\dots,\alpha^{(M)}\}$ with the $i^{th}$ vector having a size of $N_{i}$. In order to devise a distributed algorithm, the SVM optimization problem is relaxed by substituting the constraint $\sum_{i=1}^{n}\alpha_{i} y_{i}=0$ by a penalty function $0.5MZ(\sum_{i=1}^{n}\alpha_{i} y_{i})$, where $Z>0$. Thus, the following constrained optimization problem is obtained:
\begin{equation}\label{eq:dist_svm_dual}
\begin{aligned}
& \underset{\alpha}{\text{maximize}}
& & F(\alpha) = \sum_{i=1}^{n}\alpha_{i} - \frac{1}{2}\sum_{i,j=1}^{n}\alpha_{i}\alpha_{j} y_{i} y_{j}k(x_{i},x_{j}) \\
& & & - \frac{MZ}{2}\bigg(\sum_{l=1}^{n}\alpha_{l} y_{l}\bigg) \\
& \text{subject to}
& & 0\leq\alpha_{i}\leq C \beta_{i}, i = 1, \ldots, n   \\
\end{aligned}
\end{equation} 
Note that the objective function $F(\alpha)$ is strictly concave in all its arguments and the constraint set is convex, compact and nonempty.

The convex optimization problem defined in \eqref{eq:dist_svm_dual} is next partitioned into $M$ sub-problems through Lagrangian decomposition. Therefore, the $e^{th}$ unit's optimization problem becomes
\begin{equation}\label{eq:dist_svm_dual_partitioned}
\begin{aligned}
& \underset{\alpha_{i}^{(e)}\in[0, C \beta_{i}]}{\text{maximize}}
& & F_{e}(\alpha) = \sum_{i\in S_{e}}\alpha_{i} - \frac{1}{2}\sum_{i\in S_{e}}\sum_{j=1}^{n}\alpha_{i}\alpha_{j} y_{i} y_{j}k(x_{i},x_{j}) \\
& & & - \frac{MZ}{2}\bigg(\sum_{i\in S_{e}}\alpha_{i} y_{i}\bigg).
\end{aligned}
\end{equation} 

The individual optimization problems of the units are interdependent. Therefore, they cannot be solved individually without exchanging information between all the processing units. We use a fusion center that collects and distributes support vectors (SVs) among the individual processing units similar to the work of Alpcan and Bauckhage \cite{alpcan2009distributed}.

\subsubsection{Theory of Local Intrinsic Dimensionality (LID)}\label{sec:lid_theory}
LID is an expansion-based measure of the intrinsic dimensionality of the underlying data subspace \cite{LID1_Houle}. Expansion models of dimensionality have previously been successfully employed in a wide range of applications, such as manifold learning, dimension reduction, similarity search and anomaly detection \cite{LID1_Houle,amsaleg2015estimating}. \emph{In this paper, we use LID to characterize the intrinsic dimensionality of regions where attacked samples lie and create a weighting mechanism that de-emphasizes the effect of samples that have a high likelihood of being adversarial examples.} Refer to \cite{LID1_Houle} for more details concerning the theory of LID. The formal definition of LID \cite{LID1_Houle} is given below.

\textbf{Definition 1} (Local Intrinsic Dimensionality).\\
\textit{
	Given a data sample $x\in X$, let $R > 0$ be a random variable denoting the distance from $x$ to
	other data samples. If the cumulative distribution function $F(r)$ of $R$ is positive and continuously
	differentiable at distance $r > 0$, the LID of $x$ at distance $r$ is given by:}
\begin{equation}\label{eq:lid} 
\text{LID}_{F}(r)\triangleq \Lim{\epsilon \rightarrow 0}\frac{\text{ln}\big(F((1+\epsilon)\cdot r)/F(r)\big)}{\text{ln}(1+\epsilon)}=\frac{r\cdot F'(r)}{F(r)},
\end{equation}
\textit{whenever the limit exists.}

The last equality of \eqref{eq:lid} follows by applying L'H\^{o}pital's rule to the limits \cite{LID1_Houle}. The local intrinsic dimension at $x$ is in turn defined as the limit, when the radius $r$ tends to zero:
\begin{equation}\label{eq:lid_2}
\text{LID}_{F}=\Lim{r \rightarrow 0}\text{LID}_{F}(r).
\end{equation}

$\text{LID}_{F}$ describes the relative rate at which its cumulative distribution function $F(r)$ increases as the distance $r$ increases from $0$. In the ideal case where the data in the vicinity of $x$ is distributed uniformly within a subspace, $\text{LID}_{F}$ equals the dimension of the subspace; however, in practice these distributions are not ideal, the manifold model of data does not perfectly apply and $\text{LID}_{F}$ is not an integer \cite{LID_sarah_ICLR}. Nevertheless, the local intrinsic dimensionality is an indicator of the dimension of the subspace containing $x$ that would best fit the data distribution in the vicinity of $x$.

\subsubsection{Estimation of LID}
Given a reference sample $x \sim \mathcal{P}$, where $\mathcal{P}$ represents the data distribution, the Maximum Likelihood Estimator of the LID at $x$ is defined as follows \cite{amsaleg2015estimating}:
\begin{equation}\label{eq:lid_estimation}
\widehat{\text{LID}}(x)=-\Bigg(\frac{1}{k}\sum_{i=1}^{k}\text{log}\frac{r_{i}(x)}{r_{\text{max}}(x)}\Bigg)^{-1}.
\end{equation}
Here, $r_{i}(x)$ denotes the distance between $x$ and its $i$-th nearest neighbor within a sample of $k$ points drawn from $\mathcal{P}$ and $r_{\text{max}}(x)$ is the maximum of the neighbor distances. The above estimation assumes that samples are drawn from a tight neighborhood, in line with its development from Extreme Value Theory. In practice, the sample set is drawn uniformly from the available training data (omitting $x$ itself), which itself is presumed to have been randomly drawn from $\mathcal{P}$. Note that the LID defined in \eqref{eq:lid_2} is the theoretical calculation and that $\widehat{\text{LID}}$ defined in \eqref{eq:lid_estimation} is its estimate.

When the training data $X$ is large, computing neighborhoods with respect to the entire dataset $X$ can be prohibitively expensive. The LID value of a sample $x$ can be estimated from its $k$-nearest neighbor set within a randomly-selected sample (\textit{mini-batch sampling}) of the dataset $X$ \cite{LID_sarah_ICLR}. Given that the mini-batch is chosen to be sufficiently large so as to ensure that the $k$-nearest neighbor sets remain in the vicinity of $x$, estimates of LID computed for $x$ within the mini-batch would resemble those computed within the full dataset $X$. Therefore, in order to reduce the computational cost, we use \textit{mini-batch sampling} in our experiments.

\subsection{Kernel LID (K-LID)}\label{sec:lid_calculation}
The section below describes the novel LID approximation that we introduce followed by the procedure used to obtain the sample weights $\beta_{i}$.

\subsubsection{Calculating LID w.r.t. labels}
As explained earlier in Section \ref{sec:lid_theory}, LID is usually concerned with the data $X$ and is independent of the corresponding labels $y$. Since only the labels are maliciously flipped in a label flipping attack, the LID values of $X$, before and after the attack would remain the same. Therefore, we devise three label dependent LID variations as follows.

Take $S:=(X,y)$ to be the full training data set with labels. Then, define $S^{j}:=(X^{j},y^{j})$, where $j=\{\pm1\}$ as the set of data samples that carry the same label $j$. 
\begin{itemize}
	\item \textbf{In-class LID}: For each $x_{l}\in X^{j}$, the LID is calculated w.r.t. $x_{h\neq l}\in X^{j}$, for $j=\{\pm1\}$. In-class LID of a particular sample gives the dimension of the subspace w.r.t. the data distribution of the same class. 
	\item \textbf{Out-class LID}: For each $x_{l}\in X^{j}$ the LID is calculated w.r.t. $\{x_{h}\in X \mid x_{h}\notin X^{j}\}$, for $j=\{\pm1\}$. Out-class LID gives the dimension of the subspace in which a particular sample lies w.r.t. the data distribution of the opposite class. 
	\item \textbf{Cross-class LID}: Define cross-class LID as the ratio between the in-class LID and the out-class LID. 
\end{itemize}
In our experiments we use \textit{cross-class LID} to highlight samples that likely to be adversarial samples.

\subsubsection{Kernel LID calculation}
For the above label dependent LID variations to give distinguishable LID distributions for attacked and benign samples, the data clouds from the two classes need to be physically separated. 
LID in its original form is calculated using the Euclidean distance (although the underlying distance measure does not necessarily have to be Euclidean \cite{LID_sarah_ICLR}) between samples in the input space $x\in\mathcal{R}^d$. As non-linearly separable datasets have data from both classes inter-weaved, the aforementioned LID values of attacked samples and benign samples would have overlapping distributions.

To have two distinguishable LID distributions for attacked and benign samples, $X$ needs to be non-linearly transformed to a high dimensional space where the data from two classes are well separated. This approach is similar to the SVM training procedure, where the data is transformed to higher dimensional spaces where the data becomes linearly separable. In SVMs, instead of explicitly transforming data to the high dimensional space and calculating the dot product between samples, the kernel trick is used to implicitly obtain the dot products (i.e., $k(x,x')=\langle\varPhi(x),\varPhi(x')\rangle$). While it is feasible when using certain kernels to obtain the transformed $\varPhi(x)$ values, for kernels such as the \textit{radial basis function kernel} (RBF) it is not possible and only the dot product can be obtained. Thus, calculating the LID in the hypothesis space using Euclidean distance is not possible. 

The RBF kernel function non-linearly transforms the squared Euclidean distance between two data points $x$ and $x'$ as $K(x,x')=\exp(-\gamma\Vert x-x' \Vert^{2})$ in order to obtain the corresponding kernel matrix entry. The RBF kernel function returns $1$ for identical data points and $0$ for dissimilar data points. As the theory of LID computes similarity using a distance function (\ref{eq:lid_estimation}), we use the reciprocal of the kernel value and propose the following variation of LID called Kernel LID (\textit{K-LID}).
\begin{equation}\label{eq:kernel_lid}
\widehat{\text{K-LID}}(x)=-\Bigg(\frac{1}{k}\sum_{i=1}^{k}\text{log}\frac{t_{i}(x)}{t_{\text{max}}(x)}\Bigg)^{-1}.
\end{equation}
Here, $t_{i}(x)$ denotes the distance between $x$ and its $i$-th nearest neighbor calculated as $t_{i}(x)=(1/K(x,x_{i}))-1$ and $t_{\text{max}}(x)$ denotes the maximum distance among the neighbor sample points. In the remainder of this paper, we will use \eqref{eq:kernel_lid} to estimate K-LID values.

The following theorem shows the relationship between K-LID values and LID values calculated using Euclidean distance in the input space. Define $\Delta$ as a Euclidean distance measure. Then, following the above transformation of distance, $t$ can be written as $t=\frac{1}{\exp^{(-\gamma \Delta^2)}}-1$.

\textbf{Theorem 1:} The LID calculated using Euclidean distance in input space (i.e., $\text{LID}_{\Delta}$) is proportional to the K-LID calculated using transformed distances (i.e., $\text{K-LID}_{t}$) as,
\begin{equation}
\text{LID}_{\Delta}=\dfrac{2\gamma\Delta^2\exp^{(\gamma\Delta^2)}}{\exp^{(\gamma\Delta^2)}-1}\times \text{K-LID}_t.
\end{equation}

\textbf{Proof:} The proof immediately follows from Theorem 3 of \cite{houle2013dimensionality}.

\begin{figure}[t]
\centering
\begin{subfigure}{0.23\textwidth} 
	\includegraphics[width=\textwidth]{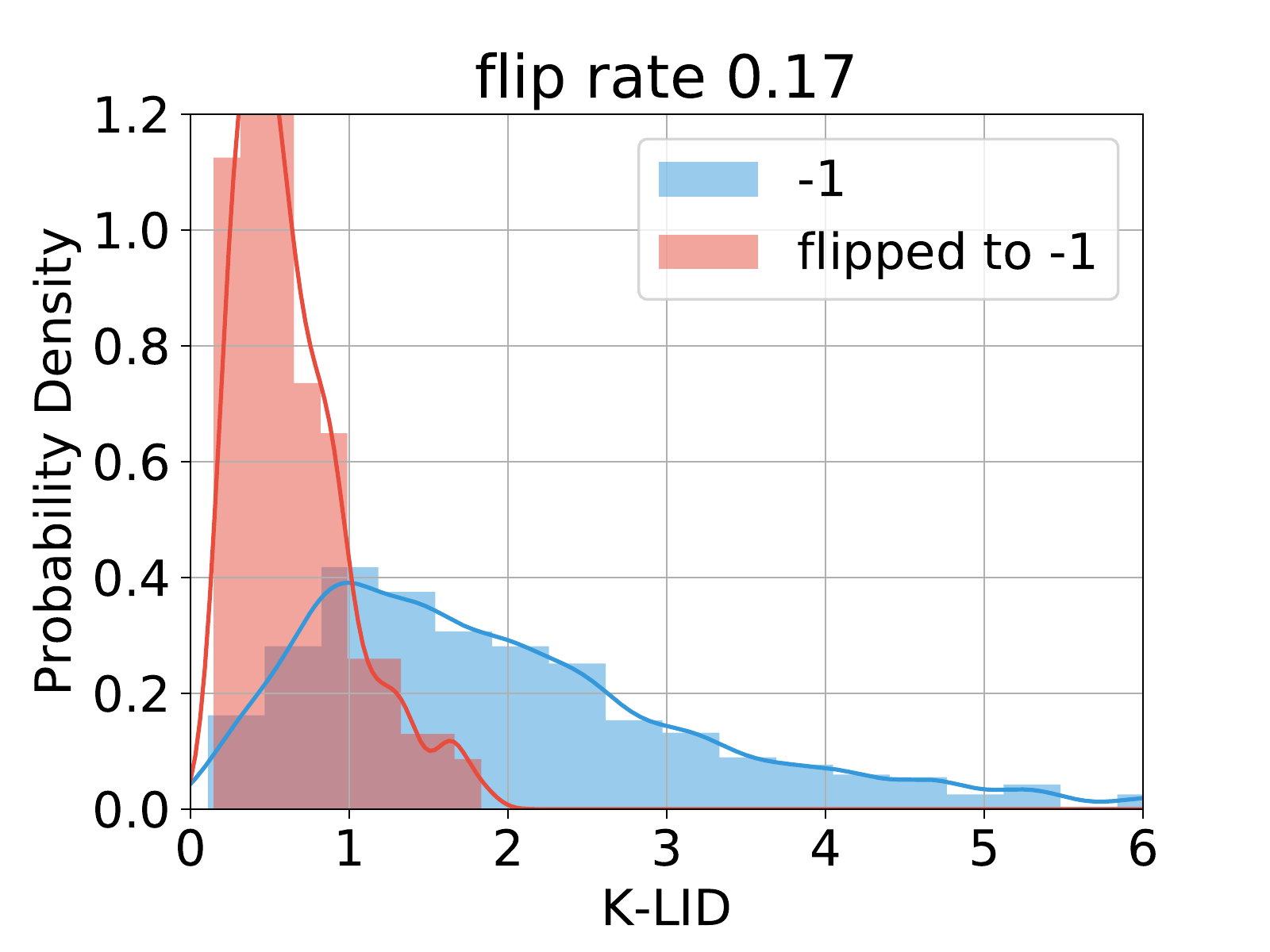}
	\caption{K-LID pdf of MNIST.} 
	\label{fig:pdf_mnist}
\end{subfigure}
\begin{subfigure}{0.23\textwidth} 
	\includegraphics[width=\textwidth]{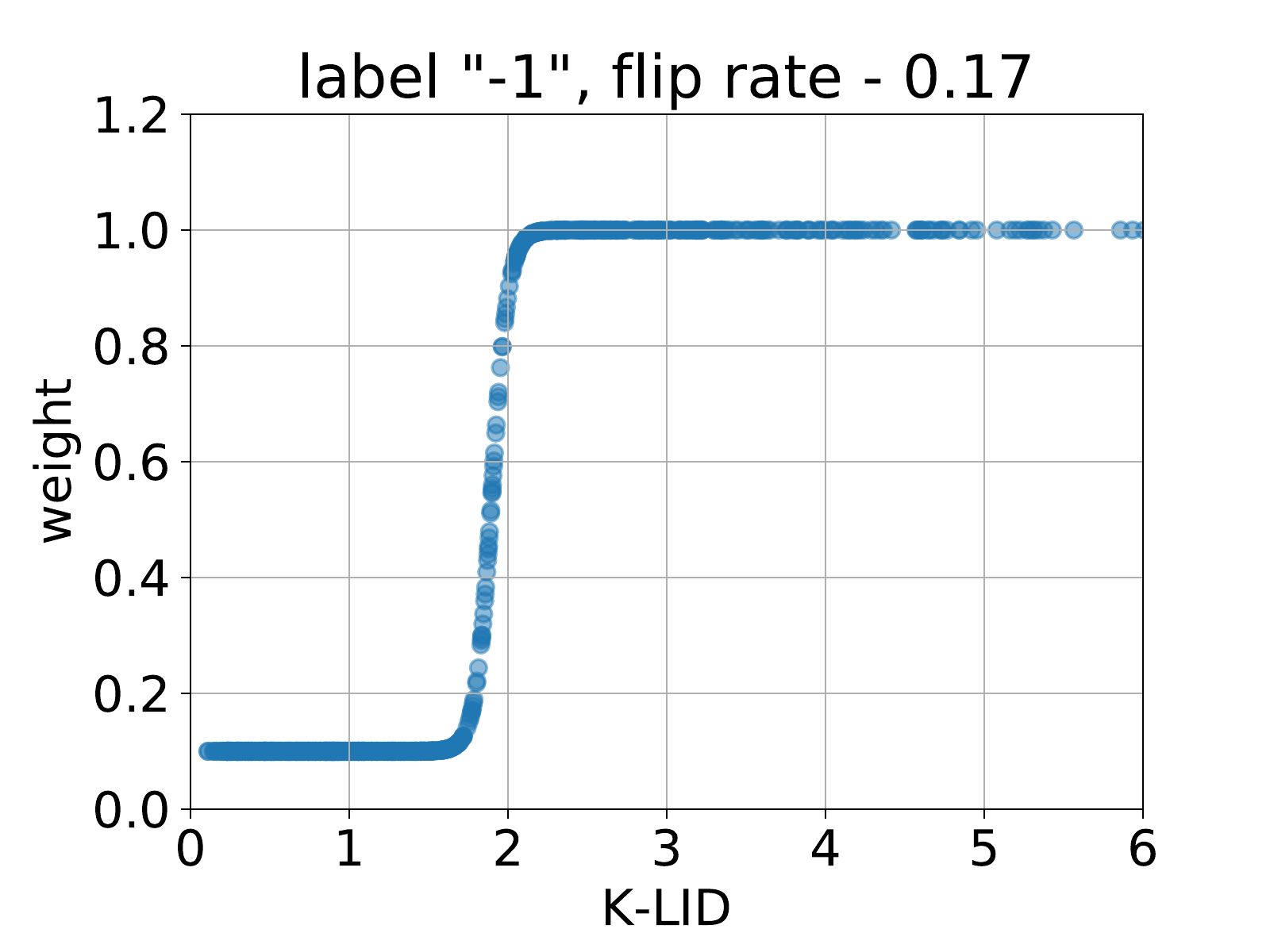}
	\caption{Weight assignment.} 
	\label{fig:weights_mnist}
\end{subfigure}
\caption{The probability density functions and the corresponding weight distributions when different $\gamma$ values are used in the K-LID calculation.}
\label{fig:klid_performance}
\end{figure}

\subsection{K-LID Distributions of Attacked and Pristine Samples}\label{sec:klid_attacked_non_attacked}
We discuss here the intuition behind using K-LID to identify adversarial samples during training. Flipping the label of a data sample would give it a different class assignment from most of its close neighbors. Computing K-LID estimates with respect to its neighborhood from within those samples that share the same class assignment (i.e., In-class K-LID) would then reveal an anomalous distribution of the local distance to these neighbors. Similarly, the out-class K-LID (K-LID calculated w.r.t. samples that have the opposite class assignment) would also give different distributions for flipped and non-flipped samples. Consequently, the Cross-class K-LID, which combines the distinguishing powers of In-class K-LID and Out-class K-LID, would have the power to distinguish flipped samples from non-flipped samples. 

Under poisoning attacks, the perturbed samples would be in close proximity to other data samples that share the same class assignment, yet not embedded within the benign data distribution. Therefore, perturbed samples would have an anomalous distribution of the local distance to these neighbors and would be highlighted by their cross-class K-LID values.

To build an SVM classifier resilient against adversarial attacks, we require the K-LID estimates of benign samples and attacked samples to have distinguishing distributions. In a black-box system, the two distributions can be obtained by simulating an attack and deliberately altering a subset of labels/data during training, by assuming the distributions based on domain knowledge or prior experience related to the specific application or by having an expert identify attacked and benign samples in a subset of the dataset.

Through our experiments, we aim to demonstrate the distinguishing capability of the novel K-LID that we introduced. To that end, we use the following grey-box system to obtain the K-LID distributions of attacked and benign samples while restating it can be converted to a black-box system by using any of the aforementioned procedures. Note that the learner \textit{does not} need to be aware of the type of attack being used, it only needs the K-LID distributions of attacked and benign samples.

First, to identify a suitable hypothesis space that separates the data of two classes, the learner performs an exhaustive search for the ideal RBF gamma parameter ($\gamma^{*}$) by calculating the cross-class K-LID values for the training dataset under different $\gamma^{*}$ values. \textit{Note that this $\gamma^{*}$ is used for obtaining K-LID values and is independent of the SVM training process}. The learner estimates the density functions of K-LID estimates for benign samples and attacked samples of each class under each $\gamma^{*}$ value using kernel density estimation methods \cite{silverman2018density}. Subsequently, the learner uses the Kullback-Leibler divergence (or colloquially KL distance) between the K-LID densities of benign and attacked samples as the measure of goodness of each transformation and selects the $\gamma^{*}$ value that yields the highest KL distance. The Kullback-Leibler divergence calculates the statistical distance between two probability distributions $P$ and $Q$ as $D_{KL}(P||Q) = \sum_{i} P(i)\log(P(i)/Q(i))$.

As Figure \ref{fig:pdf_mnist} shows, K-LID is a powerful metric that can give two distinguishable distributions for attacked and benign samples. Although, ideally we like to see no overlap between the two distributions, in real-world datasets we see some percentage of overlap. Figure \ref{fig:weights_mnist} shows the weight assignment function which gives weights to samples based on their K-LID values. In the following section, we explain how the weight assignment function is derived from the two K-LID distributions shown in Figure \ref{fig:pdf_mnist}.

\subsubsection{Sample Weighting Scheme}\label{sec:lid_svm_label_flipping}
Define $p^{j}_{n}$ and $p^{j}_{f}$ as the probability density functions of the K-LID values of benign samples and attacked samples, respectively, for the two classes $j=\{\pm1\}$. We assume there are two possible hypotheses, $H^{j}_{n}$ and $H^{j}_{f}$. For a given data sample $x'$ with the K-LID estimate $\text{K-LID}(x')$. We may write the two hypotheses for $j=\{\pm1\}$ as
\begin{equation}\label{eq:hypothesis}
\begin{aligned}
& H^{j}_{n}:\text{K-LID}(x')\sim p^{j}_{n},\\
& H^{j}_{f}:\text{K-LID}(x')\sim p^{j}_{f},
\end{aligned}
\end{equation}
where the notation ``$\text{K-LID}(x')\sim p$'' denotes the condition ``$\text{K-LID}(x')$ is from the distribution $p$''. To clarify, $H^{j}_{n}$ is the hypothesis that $\text{K-LID}(x')$ is from the distribution $p^{j}_{n}$ and $H^{j}_{f}$ is the hypothesis that $\text{K-LID}(x')$ is from the distribution $p^{j}_{f}$. The \textit{likelihood ratio} (LR) is usually used in statistics to compare the goodness of fit of two statistical models. For example, the likelihood ratio of a data sample $x'$ with the K-LID estimate $\text{K-LID}(x')$ is defined as
\begin{equation}\label{eq:likelihood_ratio}
\Lambda^{j}(\text{K-LID}(x'))={p^{j}_{n}\big(\text{K-LID}(x')\big)}/{p^{j}_{f}\big(\text{K-LID}(x')\big)}.
\end{equation}
$p(\text{K-LID}(x'))$ denotes the probability value corresponding to the K-LID of sample $x$ from the probability distribution p. To clarify, $\Lambda^{j}(\text{K-LID}(x'))$ expresses how many times more likely it is that \text{K-LID}(x') is under the K-LID distribution of benign samples than the K-LID distribution of attacked samples. As the likelihood ratio value is unbounded above, we clip the LR values of all the samples at some dataset dependent threshold after obtaining the LR values for the entire dataset.

As there is a high possibility for $p^{j}_{n}$ and $p^{j}_{f}$ to have an overlapping region, there is a risk of overemphasizing (giving a higher weight to) attacked samples. To mitigate that risk, we only de-emphasize samples that are suspected to be attacked (i.e., low LR values). Therefore, we transform the LR values such that $\Lambda^{j}(\text{K-LID}(x))\in(0.1,1)$ for $j=\{\pm1\}$. We set the lower bound for the weights to an arbitrary value close to $0$ ($0.1$ in this case) as the weighted SVM would simply disregard any training points that carry the weight $0$. The upper bound is set to $1$ to prevent overemphasizing samples. Subsequently, we fit a hyperbolic tangent function to the transformed LR values in the form of $0.55 -0.45\times\tanh(az - b)$ and obtain suitable values for the parameters $a$ and $b$. The scalar values $0.45$ and $0.55$ maintain the scale and vertical position of the function between $0.1$ and $1$. Finally, we use the K-LID value of each $x_{i}$ and use \eqref{eq:function_fitting} to obtain the corresponding weight  $\beta_{i}$.
\begin{equation}\label{eq:function_fitting}
\beta_{i}=0.55 -0.45\times\tanh(a\text{K-LID}(x_{i}) - b). 
\end{equation}

The high level procedure used to construct the K-LID-SVM under a label flipping attack is formalized in Algorithm \ref{algo:lid_svm}.
\begin{algorithm}[t]
	{\small
		\caption{K-LID-SVM Defense Algorithm}\label{algo:lid_svm}
		\begin{algorithmic}[1]
			\State \textbf{input} $\tilde{S}=(X,\tilde{y})$\Comment{contaminated training data} 
			\State \textbf{output} SVM weights $\beta$
			\For{$j=\{+1,-1\}$}\Comment{for each class}
			\State initialize $\text{K-LID}_{f},\text{K-LID}_{n},\Lambda^{j},\beta^{j}$
			\State $\gamma^{*} \gets$ \text{search for gamma that best separates the two classes} 
			\State $\text{K-LID}_{f} \gets{{\text{k-lid-calc}}(X,\tilde{y},\gamma^{*})}$\Comment{k-lid of flipped data} 
			\State $\text{K-LID}_{n} \gets{{\text{k-lid-calc}}(X,\tilde{y},\gamma^{*})}$\Comment{k-lid of non-flipped data} 
			\State $\Lambda^{j} \gets\text{calculate likelihood ratio for each K-LID value}$
			\State fit function $g(z)=0.55 -0.45\times\tanh(az - b)$ to $\Lambda^{j}$
			\State $\beta^{j} \gets g(\text{K-LID}_{X})$\Comment{obtain weight of $x_{i}\in X$ with $\tilde{y}_{i}=j$ using the K-LID of $x$} 
			\EndFor
			\State $\beta\gets \beta^{+1}~\text{and}~\beta^{-1}$
			\State K-LID-SVM = train classifier using$(X,\tilde{y},\beta)$
		\end{algorithmic}}
\end{algorithm}

\section{Experimental Results and Discussion}\label{sec:results}
The following section describes how the datasets are obtained, pre-processed and other procedures of the experimental setup. Our objective is to extensively investigate how the performance of K-LID-SVM holds against an increasing fraction of attacked training data, for each of the proposed attacks described in Section \ref{sec:attack_model}. To achieve these objectives we use several real-world categorical datasets as well as network simulation data. Our code is available at \url{https://github.com/sandamal/lid-svm}.

\subsection{Experimental Setup}
\subsubsection{Case Study: Identifying Malicious Transmission Sources}
\begin{figure}[htp]
	\centering
	{\includegraphics[width=.6\columnwidth]{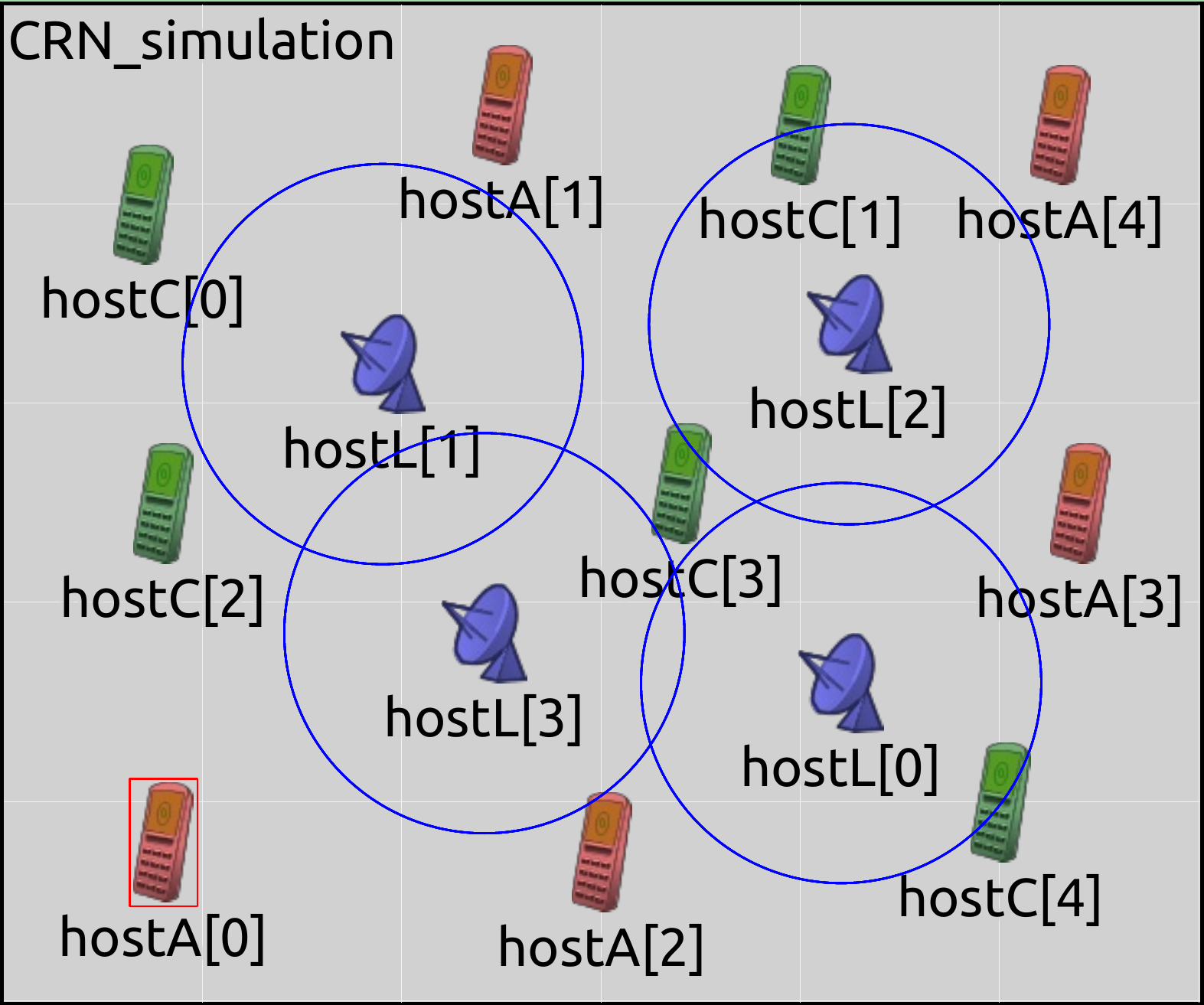}}
	\caption{A representation of the SDR listeners (blue), civilians (green) and malicious sources (red) in OMNeT++ simulator.}
	\label{fig:omnet}
\end{figure}


In recent years, there has been an increasing amount of literature on machine learning and decision making on wireless networks \cite{rajasegarar2010centered, flowers2019evaluating}. What follows is an experimental evaluation of the developed defense scheme (using DSVM) in the context of a Software-defined radio (SDR) based surveillance system. SDRs with computing capabilities can serve as a low-cost scanner array that uses a distributed SVM to identify malicious transmission sources from background radio traffic in an area (Figure \ref{fig:omnet}). Due to the prevalence of encryption methods, identification of the transmission sources has to be based on their statistical characteristics. For a network with a large number of SDR listeners, a distributed classifier is the obvious choice due to being scalable and efficient as it requires less communication overhead.

To obtain data, we use the INET framework for OMNeT++ \cite{Varga:2008:OOS:1416222.1416290} to simulate the actions of the transmitters, receivers and listeners (OMNeT simulation data available at \url{https://github.com/sandamal/omnet_simulation}). During the initial stage of system deployment, the SDR nodes collect data to train the classification models. Subsequently, during the operational stage, the SDR nodes would use the learned models to identify new transmission sources in the area. During the data collection phase and an attacker could gain access to one or more of the listener nodes to carry out a label flipping attack. This may happen through the use of malware or unauthorized access. For poisoning attacks, however, the attack can be carried out by altering the communication parameters of the malicious transmission sources.

For more realistic simulation, we consider signal attenuation, signal interference, background noise and limited radio ranges. The nodes (civilians, malicious sources and listeners) are placed randomly within the given confined area. Due to the random placement, some transmission nodes can be outside the listening range of any of the listeners. It is also possible for several transmission sources to be placed in close proximity, thereby creating signal interference. However, this is reflective of a real-world scenario where it is not possible to know beforehand where the transmission sources are positioned in a given area.

The simulator allows control of the frequencies and bit rates of the transmitter radios, their communication ranges, interference ranges, message sending intervals, message lengths, the sensitivity of the receivers and minimum energy detection of receivers among other parameters. Note that for simplicity, we assume that the listener nodes are wideband receivers, which allows them to capture data on all possible channels. However, transmission sources have fixed channels. We assume that all nodes communicate securely, therefore the listeners are unable to access the content of the captured messages. Following the simulations, we extract the following features from the data received by the listener nodes to classify transmission sources (as a civilian or a malicious source):
\begin{itemize}
	\item Duration of reception
	\item Message length
	\item Inter arrival time (IAT)
	\item Carrier frequency
	\item Bandwidth
	\item Bitrate
\end{itemize}


The duration, message length and IAT of the messages received by the listener during an hour is averaged every five minutes, which results in $36$ ($12\times 3$) features in total. Adding the latter three features (fixed for each transmission source) gives the full feature vector of $39$ features. The collected data is then fed to an attack algorithm. We consider a cellular-like architecture with $M=5$ listening nodes per region. Therefore the attacked data is then randomly divided among the listener nodes. Due to the random division of attacked training data, there is a high probability for all the compute nodes to be affected by the attack. Therefore, our approach considers the worst-case scenario for the learning system.

\subsubsection{Benchmark Datasets}
We also evaluate the effectiveness of K-LID-SVM on four real-world datasets used in \cite{alfa}: Acoustic, Ijcnn1, Seismic and Splice as well as MNIST. Note that the high computational complexities of the attacks make it infeasible to be performed on larger datasets. We report the performance of K-LID-SVM using the error rate (i.e., percentage of samples wrongly classified) on a separate test set, using 5-fold cross-validation. The average error rates are reported as the attack rate is increased from $0\%$ to $30\%$. Table \ref{tab:dataset_description} gives the chosen parameters and the number of samples used in each training and test set.

For each dataset, we compute the SVM hyper-parameters (i.e., $C$ and $\gamma$) using a 5-fold cross-validation process. Kernel density estimation was performed using a Gaussian kernel with bandwidth set using a cross-validation approach. The effects of kernel density estimation and clipping of the LR values at a heuristic threshold are neutralized by the smooth function fitting \eqref{eq:function_fitting}, therefore the algorithm is robust to these choices. For K-LID calculations, we tune $k$ over the range $[10,100)$ for a mini-batch size of 100, following Ma et al. \cite{LID_sarah_ICLR}.
\begin{table}[]
	\singlespacing
	\centering
	\caption{Datasets used for training and testing.}
	\label{tab:dataset_description}
	\begin{tabular}{lrrll}
		\toprule
		Dataset & Training size & Test size & $C$ & $\gamma$\\ \midrule
		MNIST    & 1,500             & 500         & 1.47   &  0.0197\\
		Acoustic & 500               & 500         & 1,024   &  0.0078\\
		Ijcnn1   & 500               & 500         & 64   &  0.1200\\
		Seismic  & 500               & 500         & 1,024   &  0.0078\\
		Splice   & 500               & 500         & 1,024   &  0.0078\\
		OMNeT    & 364               & 91         & 0.3969   &  0.7937\\
		\bottomrule
	\end{tabular}
\end{table}

\subsection{Results}
\begin{figure}[]
	\centering
	\begin{subfigure}{0.23\textwidth} 
		\includegraphics[width=\textwidth]{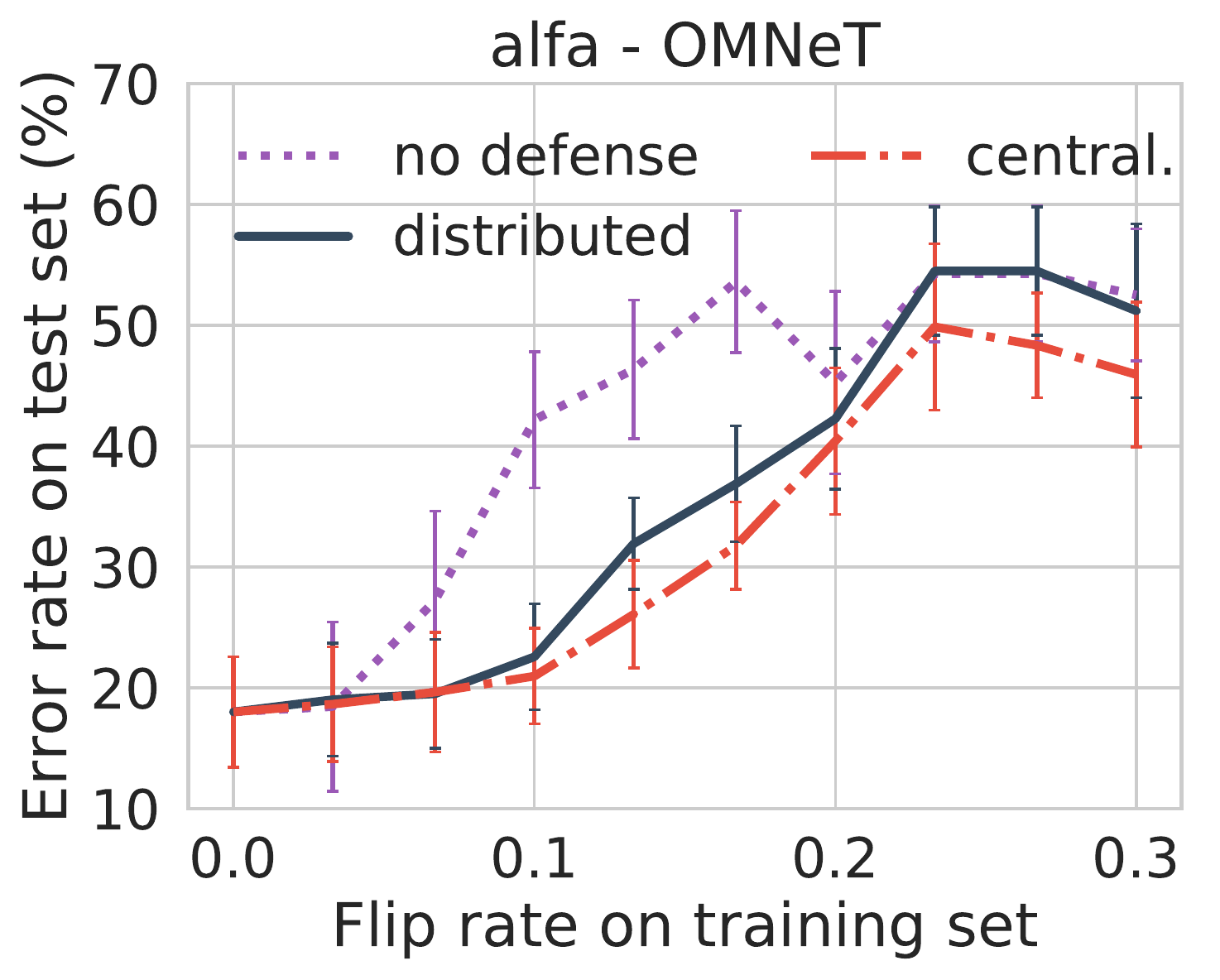}
		\caption{Test error on OMNeT data.} 
		\label{fig:omnet_alfa_distr}
	\end{subfigure}
	\begin{subfigure}{0.23\textwidth} 
		\includegraphics[width=\textwidth]{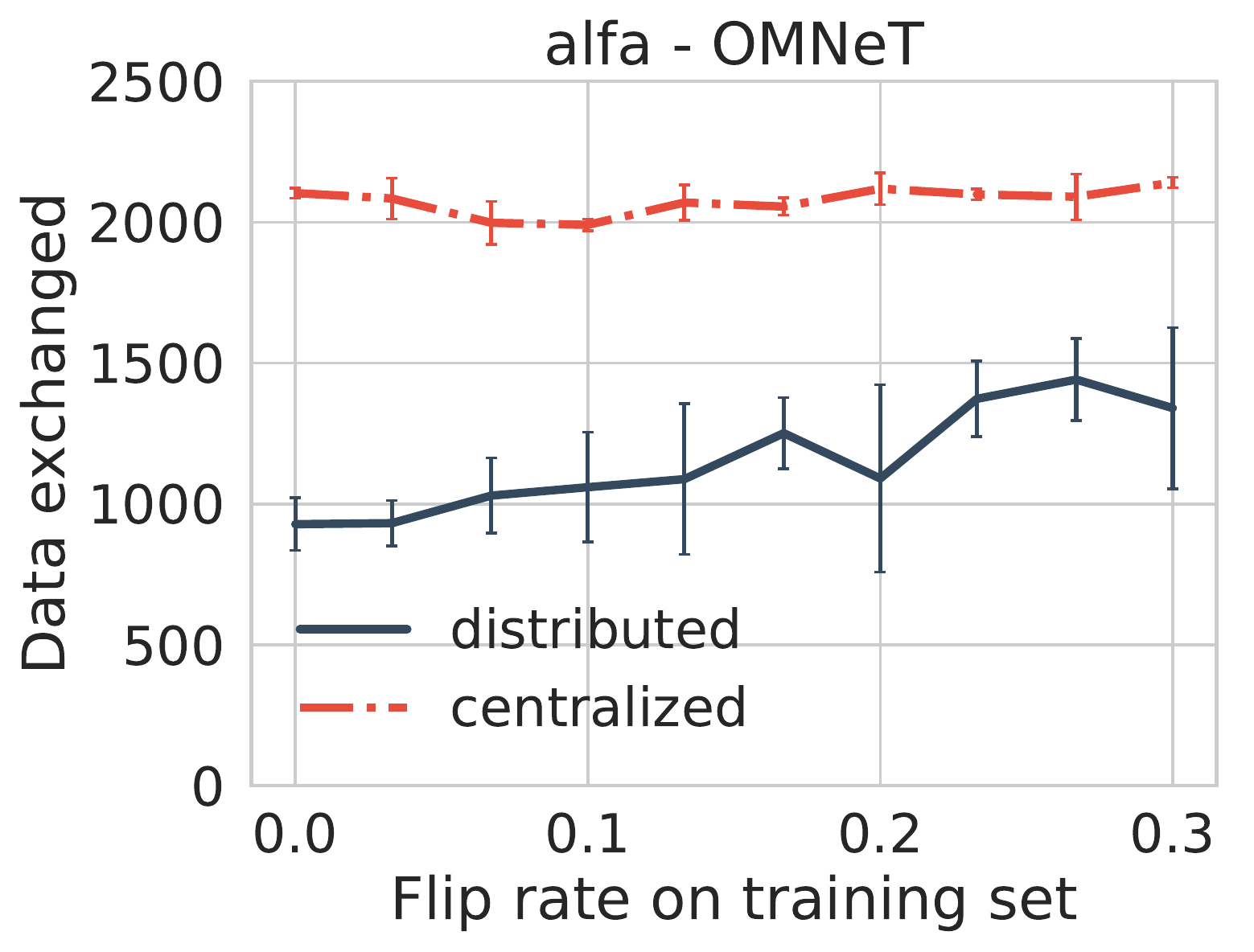}
		\caption{Data exchanged.} 
		\label{fig:omnet_SV}
	\end{subfigure}
	\caption{The classification performance of centralized K-LID-SVM and distributed K-LID-SVM and the amount of data exchanged in each case.}
	\label{fig:dist_vs_centralized}
\end{figure}
\subsubsection{Distributed detection vs. centralized detection}\label{sec:dist_vs_cent}
First, we compare the performance of a DSVM against the performance of a centralized SVM in the SDR based surveillance system. As SDRs have limited power and range, each SDR captures only the transmissions of nodes within its range. The objective of the cognitive network is for each SDR to have the ability to classify a new transmission source $x$ as a malicious source or a civilian in a distributed manner without communicating $x$ itself to the other nodes. But to obtain a classifier that can identify all the types of transmitters in the given area, the information obtained by each listener would need to be communicated to all the other SDR listeners during the training phase. This can be achieved in one of two ways: (i) have a fusion center collect all the data from the listeners and train one classification model and transfer the learned model back to the listeners (i.e., centralized solution), or (ii) have the listener SDRs (which have computing capabilities) compute classification models based on their local data, exchange the learned models with other listeners through one or more fusion centers followed by an update to their local models. For simplicity, we only consider the data exchange between the SDRs and the fusion centers (i.e., can be modeled as a distributed SVM with one fusion center). While having more than one fusion center improves the robustness of the overall system, it would increase the communication overhead due to the information exchange between the fusion centers.

Figure \ref{fig:omnet_alfa_distr} shows the error rates on the test set when malicious sources carry out alfa attacks on the distributed SDR listeners with the flip rate increasing from 0\% to 30\%. We observe that the centralized SVM solution has lower error rates compared to the DSVM solution on average. We postulate that this is due to two main reasons, (i) the optimization problem used by the DSVM \eqref{eq:dist_svm_dual} is a relaxation of the optimization problem of the centralized SVM \eqref{eq:svm_dual}, and (ii) as shown by Amsaleg et al. \cite{amsaleg2015estimating}, the MLE estimator of LID \eqref{eq:lid_estimation} is not stable on small mini-batch sizes. In the DSVM setting, each SVM node trains on the data that it receives from civilians and malicious sources within its listening range, resulting in smaller dataset sizes. Therefore, the resulting LID estimations would also be affected compared to the centralized learner.

Although the detection capability of the DSVM is less than the centralized SVM, we observe that its information exchange overhead is significantly less compared to the centralized SVM in this particular scenario. Figure \ref{fig:omnet_SV} shows the number of data points exchanged at each flip rate for the two SVM solutions. In our experiment, we consider $M=5$ listeners, therefore in the centralized SVM, the listeners would first transfer their data points to the central SVM and the SVM would transfer the SVs back to all five nodes after the SVM training process. Afterward, each node would be able to evaluate new data samples using the received SVs (note that for simplicity we are not considering the overhead of transmitting the $\alpha$ values). In the DSVM, each SVM node would use a random vector of $\alpha$ values to initialize the training process. Subsequently, after each training iteration, they would transfer their SVs to a central fusion node, which would distribute them among the other SVM nodes. Note that the DSVM iteratively trains only until the detection accuracy converges.

As Figure \ref{fig:omnet_SV} shows, the DSVM can reduce the information exchange overhead by $44.4\%$ on average with only a $3.07\%$ reduction in average detection accuracy. These findings suggest that detection using a DSVM solution is indeed the better option for the SDR based detection network.

\begin{figure}[]
	\centering
	\begin{subfigure}{0.23\textwidth} 
		\includegraphics[width=\textwidth]{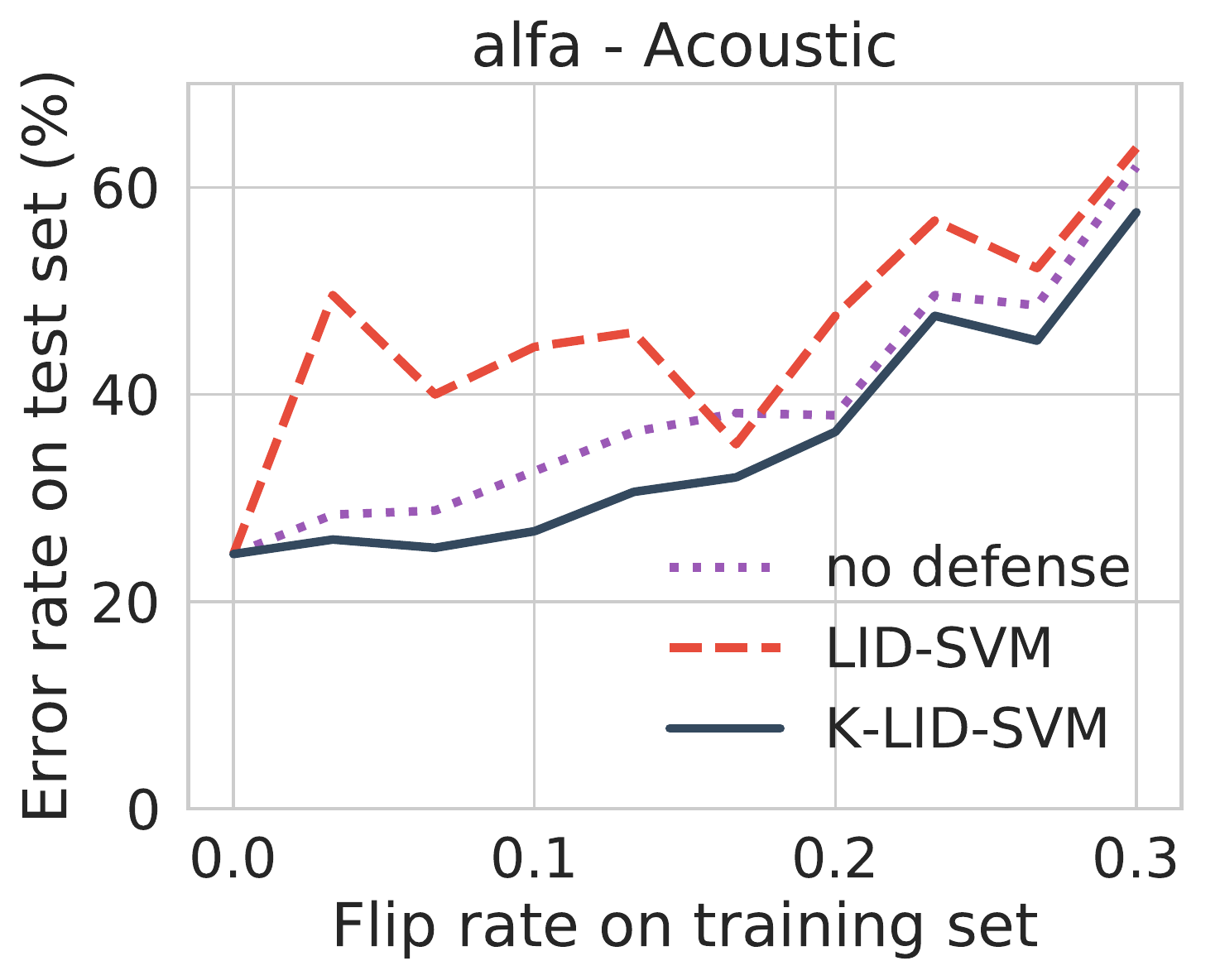}
		\caption{LID vs K-LID on acoustic.} 
		\label{fig:acoustic_klid}
	\end{subfigure}
	\begin{subfigure}{0.23\textwidth} 
		\includegraphics[width=\textwidth]{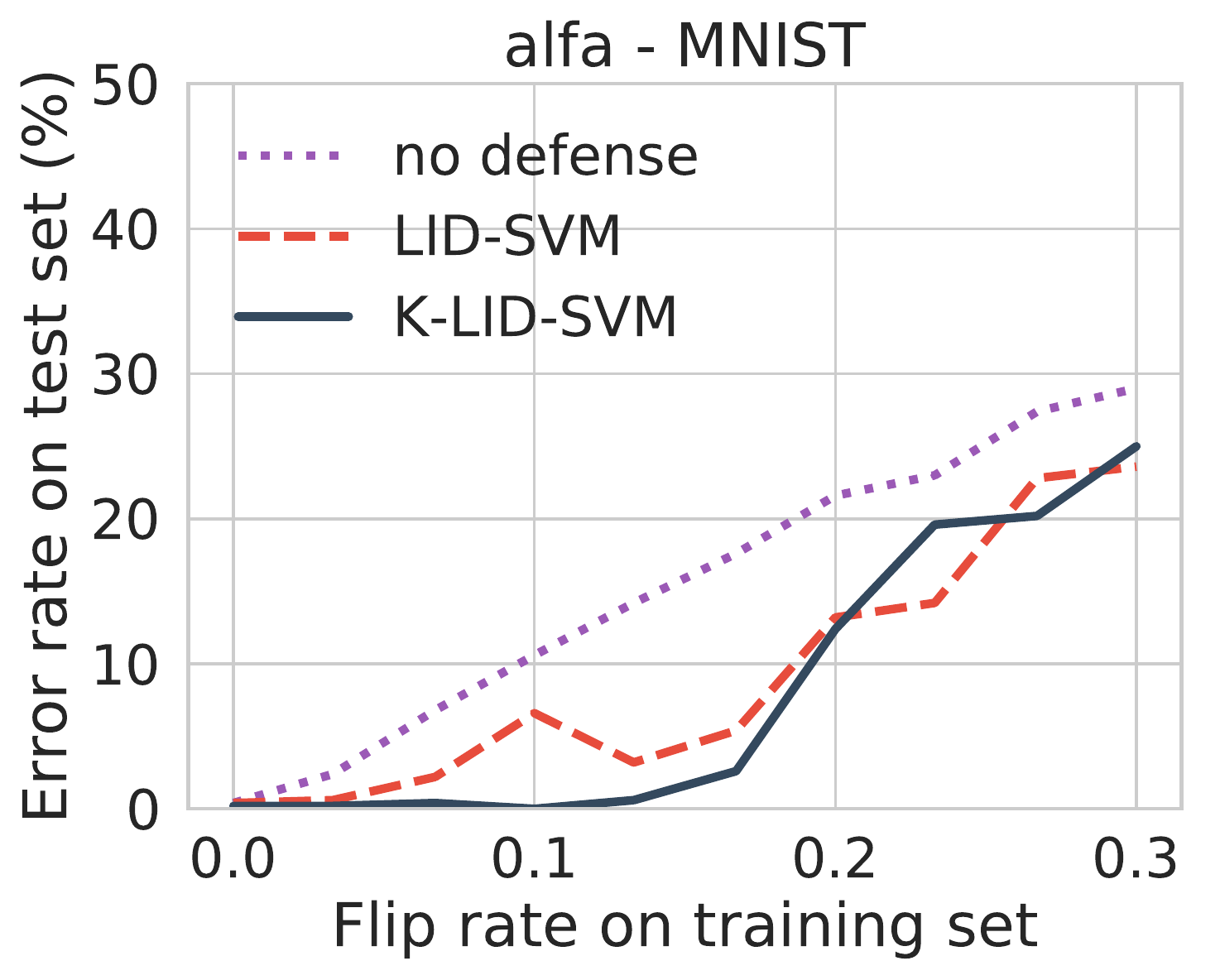}
		\caption{LID vs K-LID on MNIST.} 
		\label{fig:mnist_klid}
	\end{subfigure}
	\caption{The classification performance of LID-SVM and K-LID-SVM on two datasets.}
	\label{fig:klid_error_rates}
\end{figure}
\subsubsection{Performance of K-LID vs. LID} 
Figure \ref{fig:klid_error_rates} depicts the classification performance of K-LID-SVM vs. conventional LID-SVM. As expected, K-LID-SVM has lower error rates on average. If data from the two classes are not linearly separable in the input space, conventional LID fails to give two distinguishable distributions with a low percentage of overlap. When the percentage of overlap is high, the defense algorithm assigns a low, uniform weight to almost all samples. Such a weight assignment would make the LID-SVM equivalent to an SVM with no defense (with a sub-optimal $C$ value). Therefore, as seen in Figure \ref{fig:acoustic_klid}, LID-SVM could perform worse than an SVM with no defense in some data sets. However, K-LID, which performs the LID calculation in higher dimensional transformed spaces where the data is separated, has sufficient separation power to obtain two distinguishable distributions with a low percentage of overlap.

When using the kernel-based distance function, if the $\gamma$ used is large, two data points are considered similar only if they are close to each other in the transformed space, conversely, if $\gamma$ is small, even data points that are farther away from each other in the transformed space are considered similar. Therefore different $\gamma$ values give distinct K-LID distributions with different percentages of overlap. When there is a high percentage of overlap it is not possible to obtain meaningful weights.

\subsubsection{K-LID Distributions at High Attack Rates} 
Having a flip rate of 30\% assumes a very powerful attacker with significant influence. At high flip rates, the two distributions have a high overlap percentage and the likelihood ratio cannot be used to distinguish flipped samples from non-flipped ones, therefore the learner assigns a low, uniform weight to most samples. We believe that this increase in overlap percentage is the main reason why K-LID-SVM tends to have relatively higher error rates when the flip rate is 30\%. As explained in Section \ref{sec:lid_calculation}, we estimate K-LID using \textit{mini-batch sampling}. For a flipped sample $x_{i}$, we speculate that at a flip rate of 30\%, the probability of having other flipped samples within the randomly-selected subset increases and the K-LID calculation is influenced by other flipped samples that carry the same label $y_{i}$. A comprehensive investigation of the trade-offs among mini-batch size, K-LID estimation accuracy and detection performance is an interesting direction for future work.

\begin{table}[]
	\singlespacing
	\centering
	\caption{Average error rates across all the flip rates of each defense algorithm against the five attacks considered. The best results are indicated in \textbf{bold} font.}
	\label{tab:avg_error_rates}
	\begin{tabular}{@{}clr>{\columncolor[gray]{0.9}}rrr@{}}
		\toprule
		& Dataset  & SVM & K-LID-SVM & LS-SVM & LN-SVM \\\midrule
		{\parbox[t]{2mm}{\multirow{5}{*}{\rotatebox[origin=l]{90}{random}}}}    & MNIST    & 0.64       & \textbf{0.62}  & 12.83          & 5.17           \\
		& Acoustic & 27.80      & \textbf{26.35} & 28.14          & 28.58          \\
		& Ijcnn1   & 16.79      & \textbf{15.51} & 16.36          & 19.60          \\
		& Seismic  & 25.45      & 24.79          & \textbf{23.21} & 36.97          \\
		& Splice   & 24.77      & \textbf{20.28} & 24.07          & 30.42          \\
		& OMNeT    & 29.19 		& \textbf{24.10}   		 & 27.85  		  & 31.08  		   \\\midrule
		{\parbox[t]{2mm}{\multirow{5}{*}{\rotatebox[origin=c]{90}{farfirst}}}}  & MNIST    & 12.93      & \textbf{3.08}  & 16.72          & 18.54          \\
		& Acoustic & 39.22      & \textbf{35.01} & 39.96          & 42.14          \\
		& Ijcnn1   & 34.42      & 29.91          & \textbf{22.49} & 38.21          \\
		& Seismic  & 33.16      & \textbf{31.20} & 31.48          & 35.00          \\
		& Splice   & 31.60      & \textbf{28.29} & 31.10          & 33.21          \\
		& OMNeT    & 36.75 		& 32.10   		 & \textbf{26.18}  		  & 34.88  		   \\\midrule
		{\parbox[t]{2mm}{\multirow{5}{*}{\rotatebox[origin=c]{90}{nearest}}}}   & MNIST    & 6.73       & \textbf{1.56}  & 10.25          & 7.18           \\
		& Acoustic & 26.76      & 25.94          & 26.69          & \textbf{25.23} \\
		& Ijcnn1   & 13.90      & \textbf{12.38} & 14.66          & 18.54          \\
		& Seismic  & 20.68      & 19.82          & \textbf{18.97} & 45.62          \\
		& Splice   & 22.79      & \textbf{21.30} & 22.50          & 39.11          \\
		& OMNeT    & 30.64 	 	& \textbf{26.42}   	 	 & 31.08    	  & 37.56  		   \\\midrule
		{\parbox[t]{2mm}{\multirow{5}{*}{\rotatebox[origin=c]{90}{alfa}}}}		& MNIST    & 14.34      & \textbf{8.79}  & 16.76          & 17.00          \\
		& Acoustic & 39.94      & \textbf{36.87} & 40.25          & 42.23          \\
		& Ijcnn1   & 32.02      & 29.01          & \textbf{23.96} & 35.08          \\
		& Seismic  & 30.85      & \textbf{29.60} & 30.31          & 33.42          \\
		& Splice   & 29.90      & \textbf{26.80} & 29.40          & 34.64          \\
		& OMNeT    & 41.23 		& 31.98   		 & \textbf{25.54}  		  & 33.41  		   \\\midrule
		{\parbox[t]{2mm}{\multirow{5}{*}{\rotatebox[origin=c]{90}{alfa-tilt}}}} & MNIST    & 17.13      & \textbf{6.64}  & 16.76          & 17.00          \\
		& Acoustic & 43.87      & \textbf{40.72} & 43.78          & 43.03          \\
		& Ijcnn1   & 32.60      & 27.05          & \textbf{26.00} & 36.47          \\
		& Seismic  & 33.93      & \textbf{30.26} & 31.90          & 46.21          \\
		& Splice   & 30.47      & \textbf{26.64} & 30.58          & 43.28     	   \\
		& OMNeT    & 42.79 		& 37.68   		 & \textbf{22.81}  		  & 33.41   	   \\
		\bottomrule
	\end{tabular}
\end{table}

\subsubsection{Under Label Flipping Attacks}
We compare the performance of K-LID-SVM against LS-SVM \cite{suykens1999least} and LN-SVM \cite{biggio2011support} which have been shown to be effective against label flipping attacks (see Section \ref{sec:lit_review}). Table \ref{tab:avg_error_rates} gives the error rate of each defense mechanism averaged over all the flip rates considered ($0\%$ - $30\%$). Due to space limitations, graphs of only some experiments are shown.

\begin{figure*}[t!]
	\centering
	\begin{subfigure}{0.3\textwidth} 
		\includegraphics[width=\textwidth]{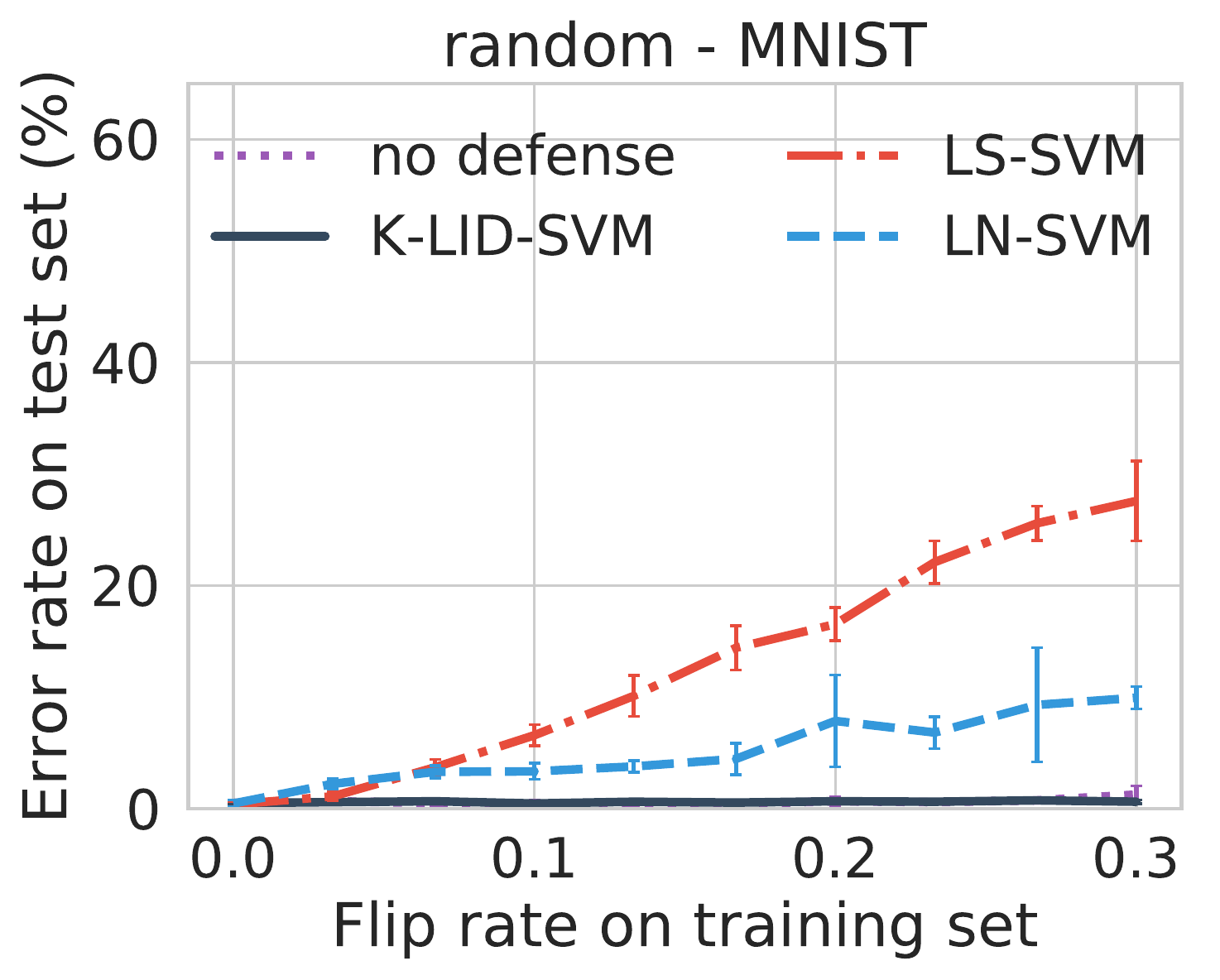}
		\label{fig:mnist_random}
	\end{subfigure}
	\begin{subfigure}{0.3\textwidth} 
		\includegraphics[width=\textwidth]{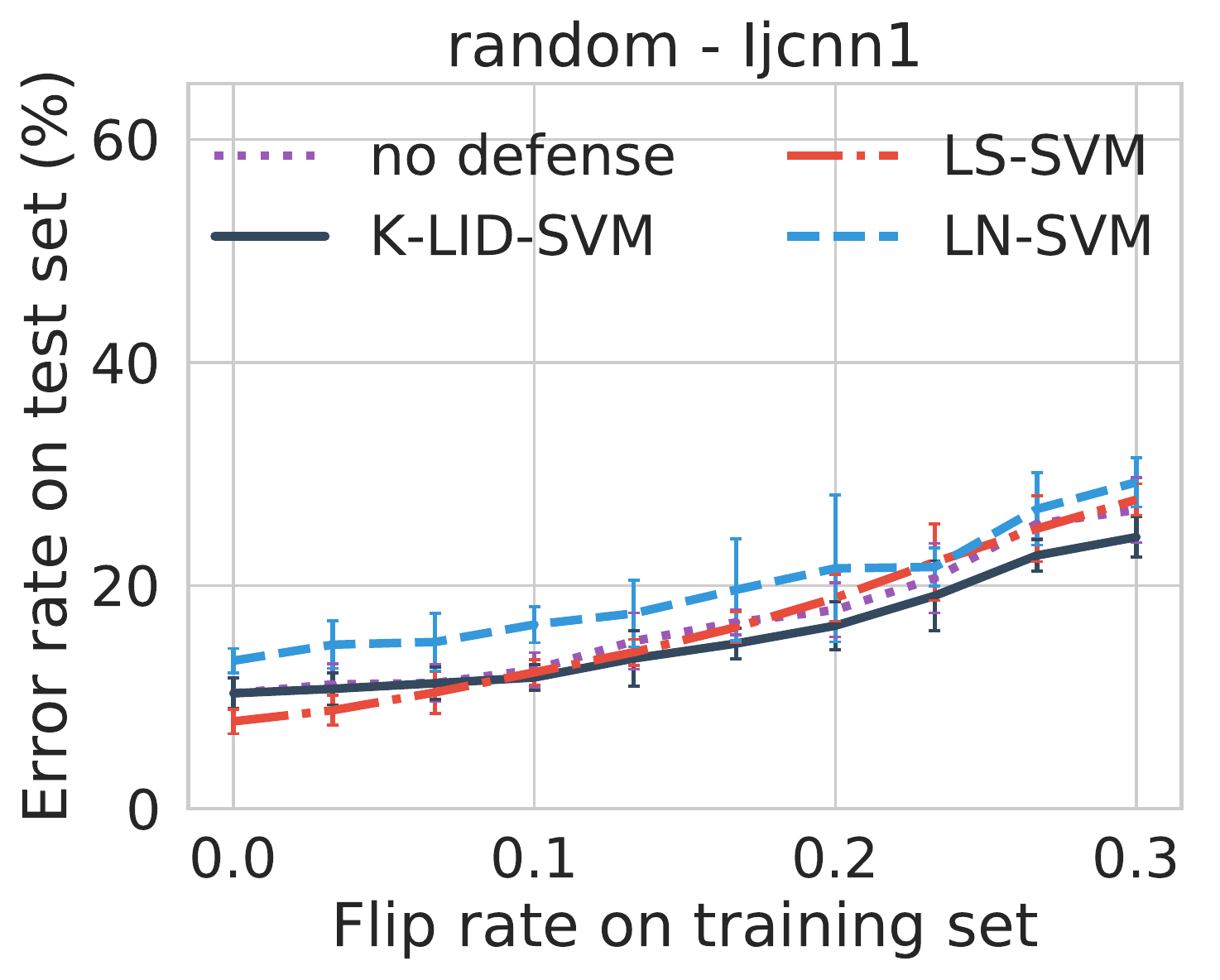}
		\label{fig:ijcnn_random}
	\end{subfigure}
	\begin{subfigure}{0.3\textwidth} 
		\includegraphics[width=\textwidth]{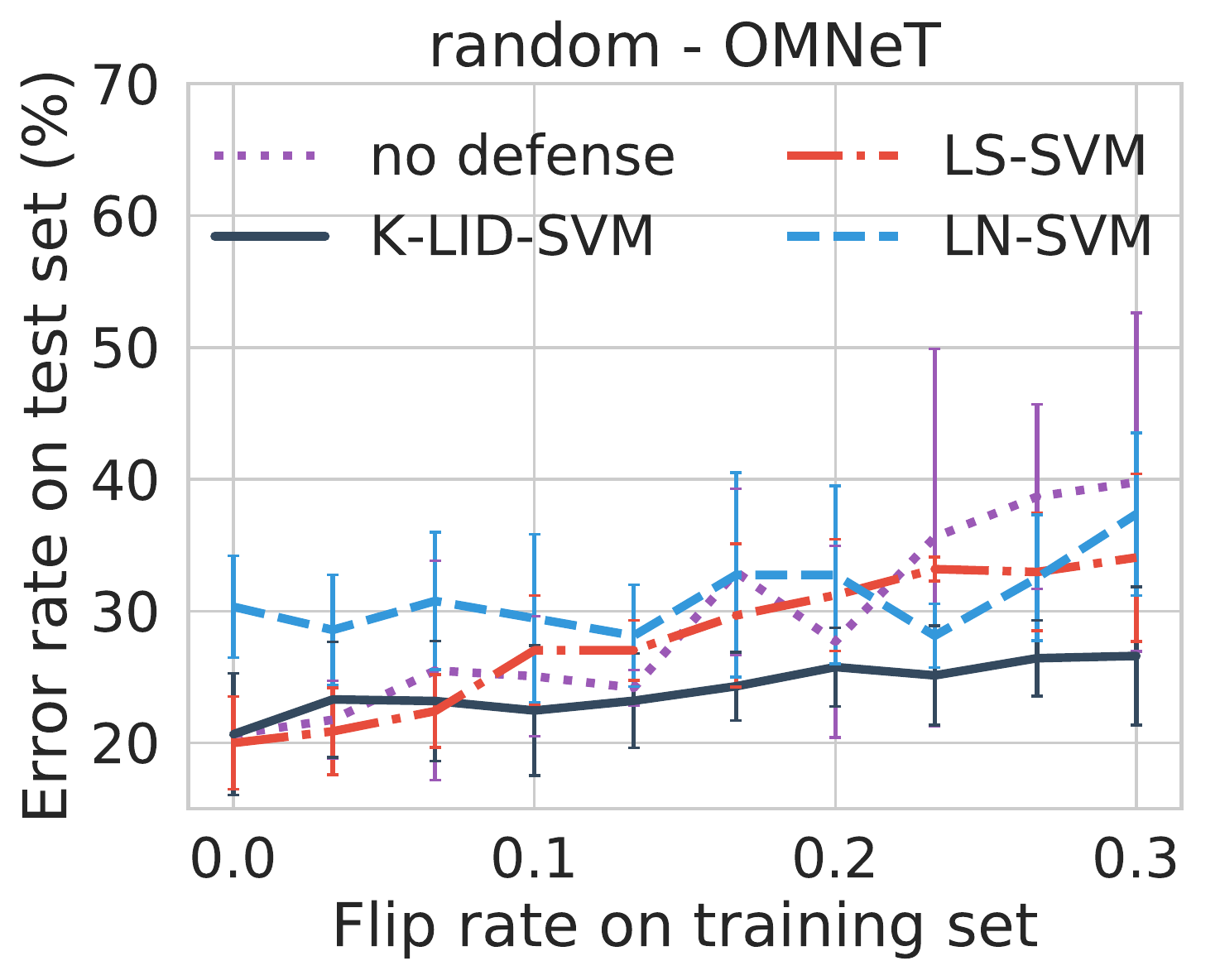}
		\label{fig:omnet_random}
	\end{subfigure}
	\caption{The average error rates of SVM, K-LID-SVM, LS-SVM and LN-SVM ($\mu=0.15$) against random label flip attacks on MNIST, Ijcnn1 and OMNeT when the training flip rate increases from 0\% to 30\%.}
	\label{fig:random_error_rates}
\end{figure*}
\textbf{Random label flips:} The performance of the binary SVM without a defense against random label flips varies from dataset to dataset. We observe that it can retain near $1\%$ error rate on MNIST, a mere $5\%$ increase in error rate on Acoustic and a $6\%$ increase on OMNeT when the flip rate is increased up to $30\%$. On Ijcnn1, Seismic and Splice however there is a $16\%$, $10\%$ and $18\%$ increase in the error rates respectively. K-LID-SVM outperforms the other defense mechanisms in all of the datasets except for Seismic where LS-SVM has a $1.58\%$ lower average error rate. On the other datasets however, K-LID-SVM has lower average error rates (up to 12\%) than the other two defenses. Figure \ref{fig:random_error_rates} depicts how the error rates vary on MNIST, Ijcnn1 and OMNeT when the training flip rate increases from 0\% to 30\%.

\begin{figure*}[t!]
	\centering
	\begin{subfigure}{0.3\textwidth} 
		\includegraphics[width=\textwidth]{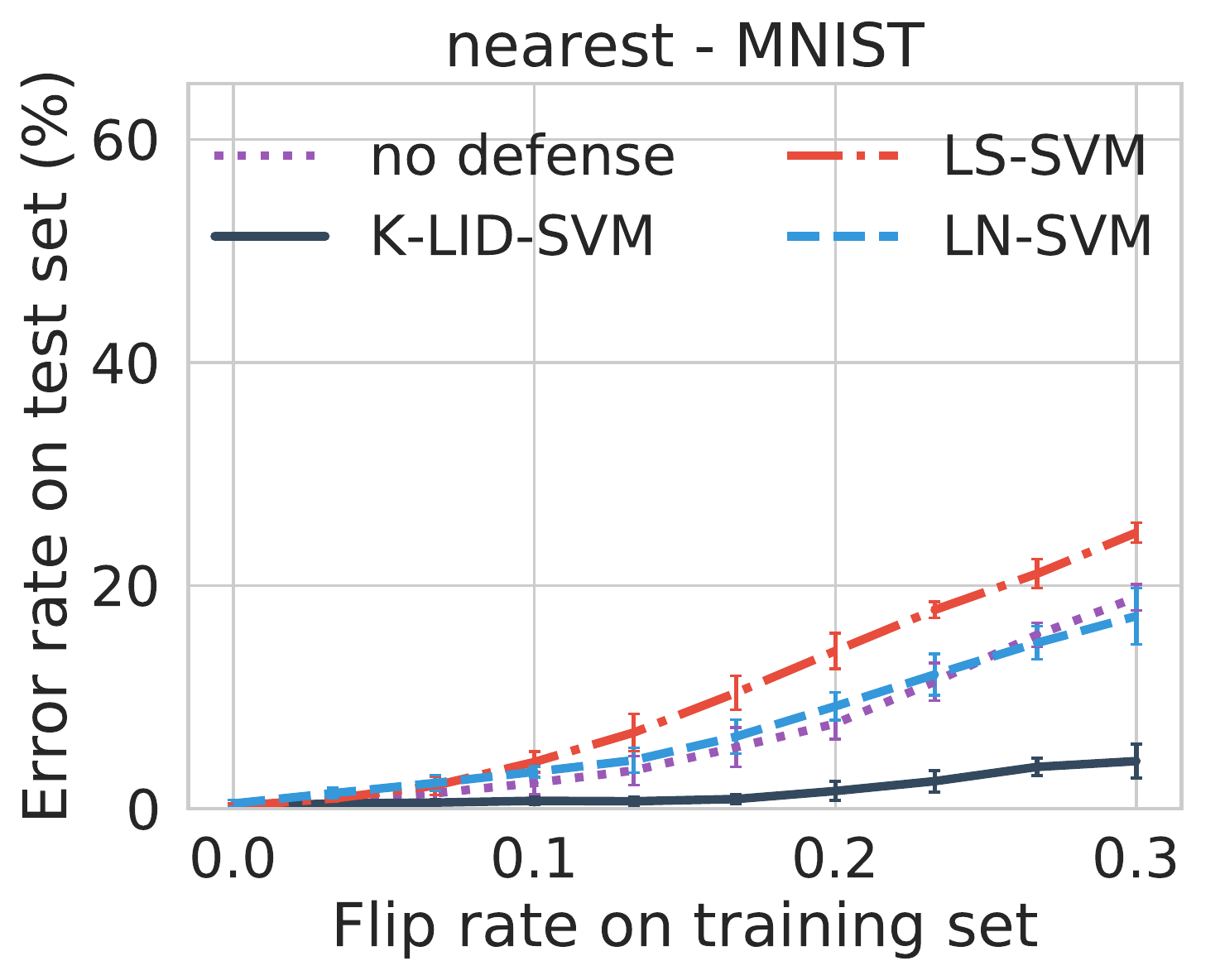}
		\label{fig:mnist_nearest}
	\end{subfigure}
	\begin{subfigure}{0.3\textwidth} 
		\includegraphics[width=\textwidth]{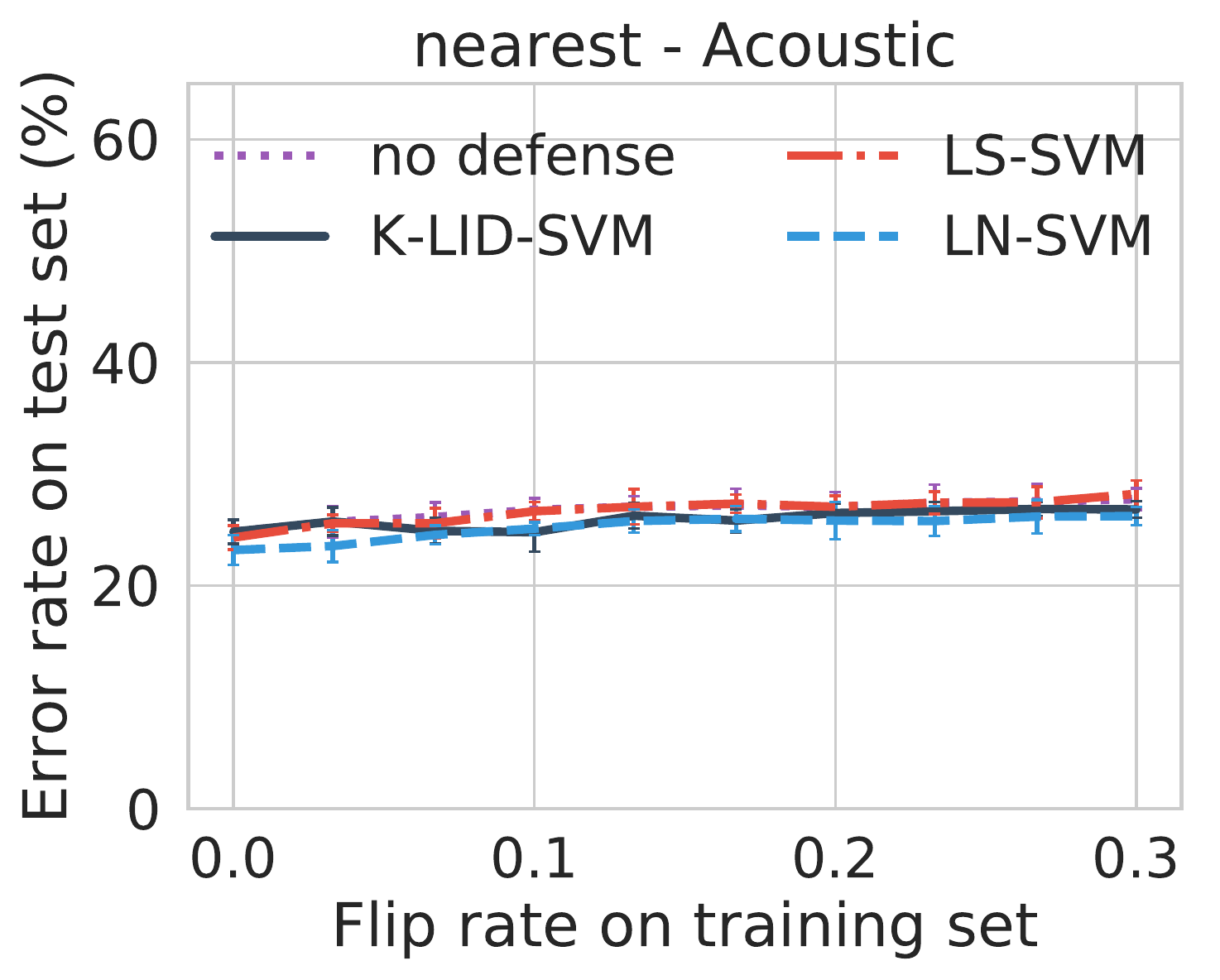}
		\label{fig:acoustic_nearest}
	\end{subfigure}
	\begin{subfigure}{0.3\textwidth} 
		\includegraphics[width=\textwidth]{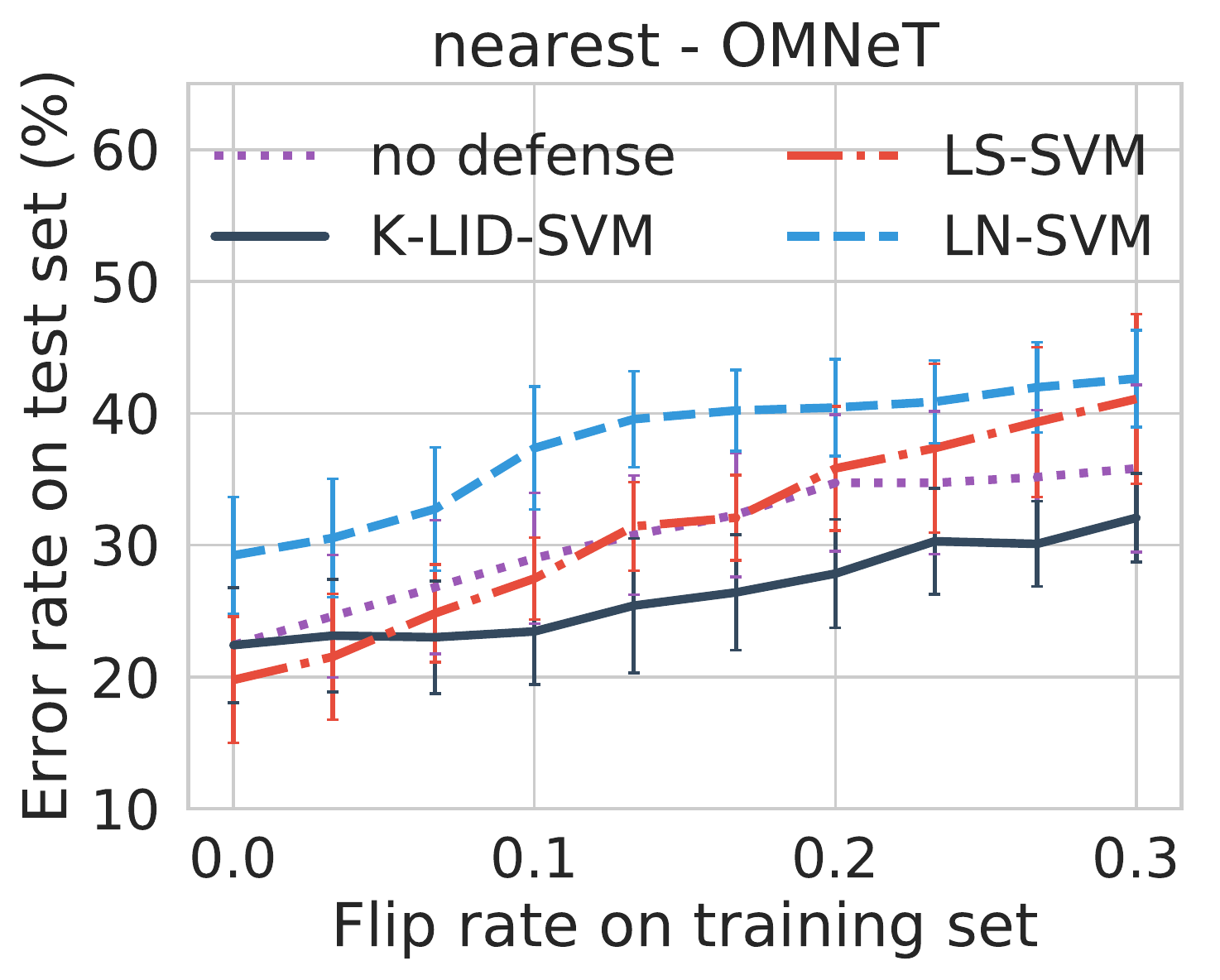}
		\label{fig:omnet_nearest}
	\end{subfigure}
	\begin{subfigure}{0.3\textwidth} 
		\includegraphics[width=\textwidth]{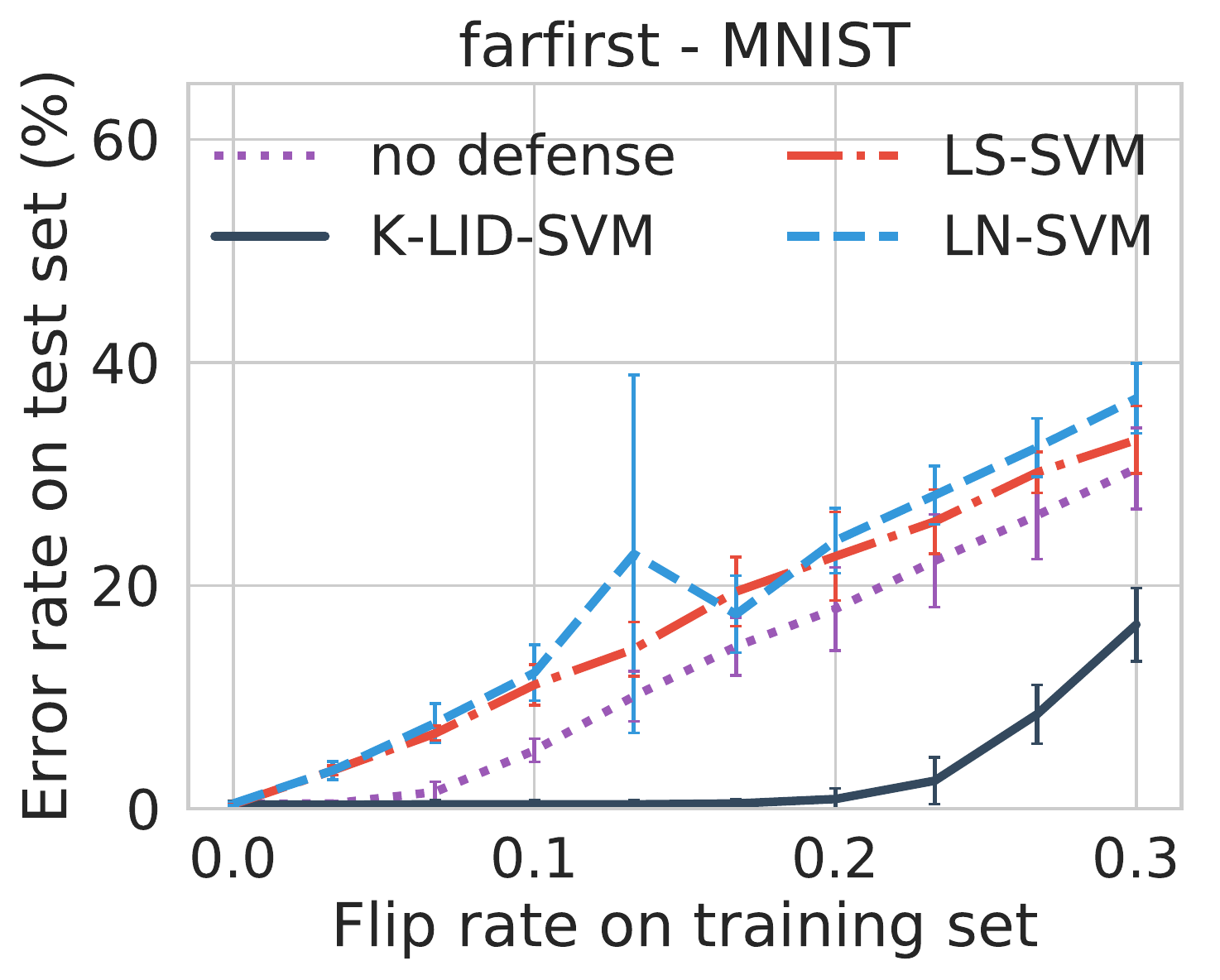}
		\label{fig:mnist_farfirst}
	\end{subfigure}
	\begin{subfigure}{0.3\textwidth} 
		\includegraphics[width=\textwidth]{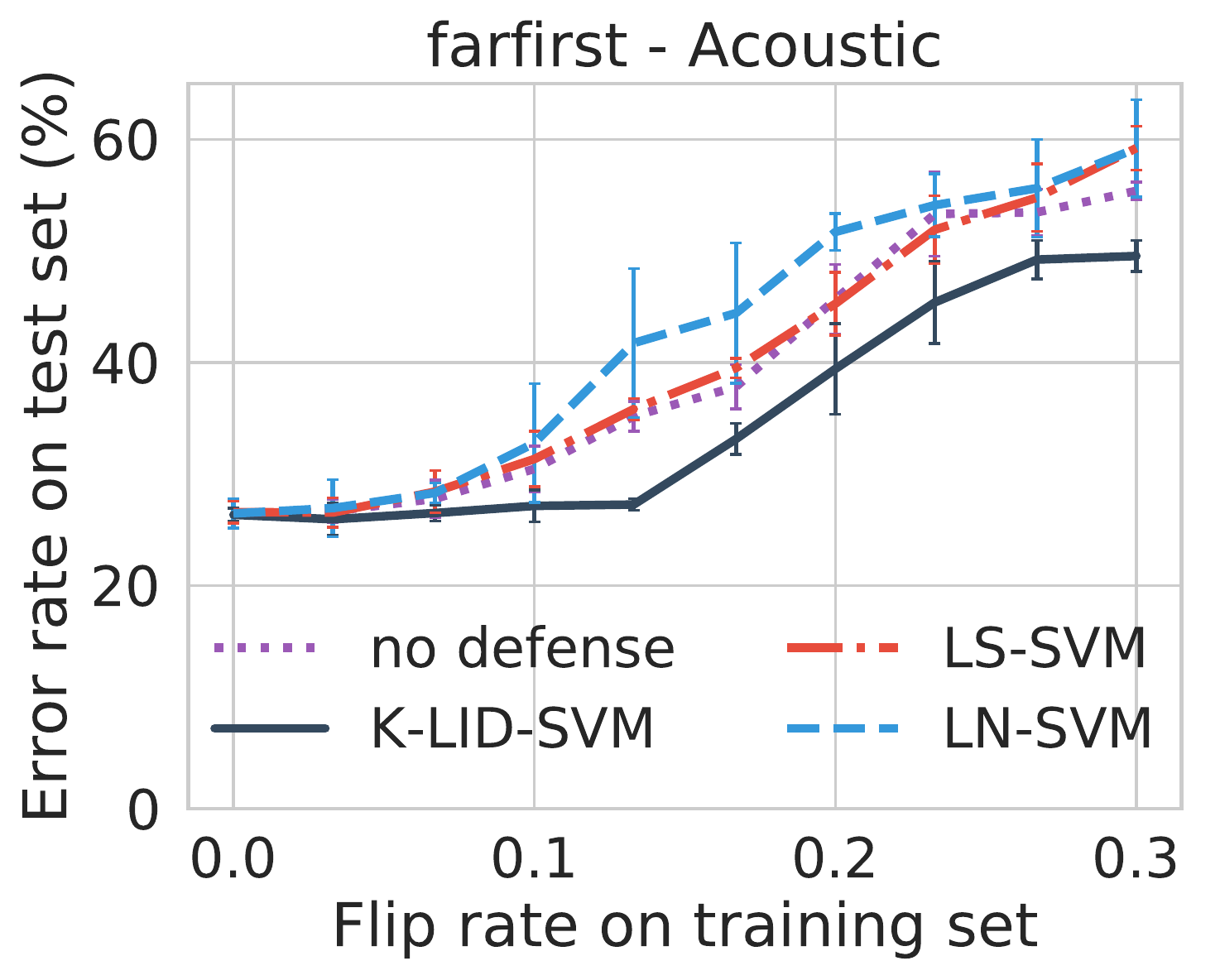}
		\label{fig:acoustic_farfirst}
	\end{subfigure}
	\begin{subfigure}{0.3\textwidth} 
		\includegraphics[width=\textwidth]{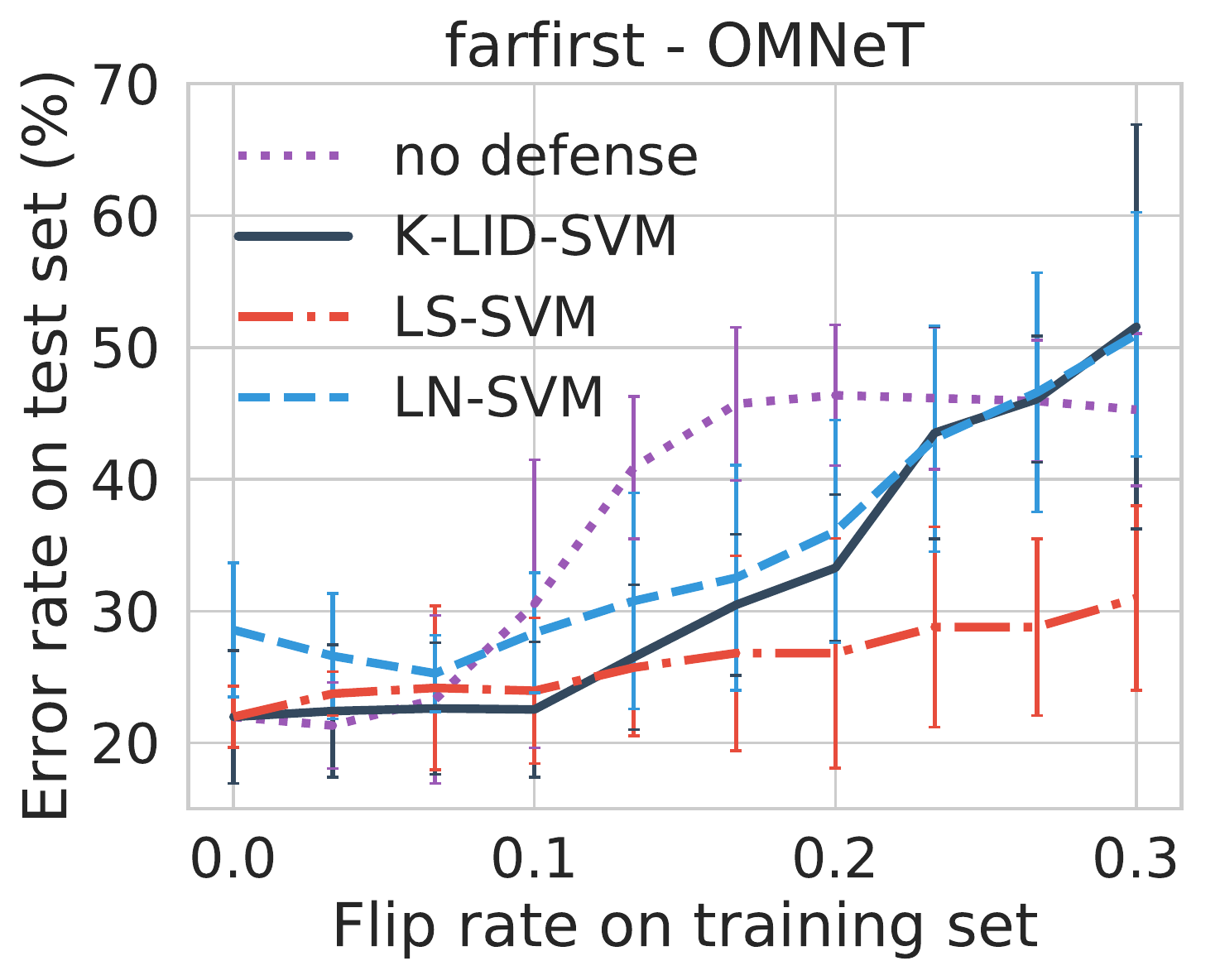}
		\label{fig:omnet_farfirst}
	\end{subfigure}
	\caption{The average error rates of SVM, K-LID-SVM, LS-SVM and LN-SVM ($\mu=0.15$) against naive label flip attacks on MNIST, Acoustic and OMNeT when the training flip rate increases from 0\% to 30\%.}
	\label{fig:naive_error_rates}
\end{figure*}
\textbf{Naive adversarial label flips:} We consider \textit{farfirst} and \textit{nearest} as naive attacks as the algorithms are relatively simpler compared to \textit{alfa} and \textit{alfa-tilt}. Although \textit{farfirst} is simple, it can have a significant impact on an undefended SVM with error rates increasing by 30\% on MNIST, 13\% on OMNeT, 29\% on Acoustic, 48\% on Ijcnn1, 27\% on Seismic and 34\% on Splice when the flip rate is increased to 30\%. In \textit{nearest}, the increase in error rates are 18\% on MNIST, 13\% on OMNeT, 3\% on Acoustic, 8\% on Ijcnn1, 1.4\% on sesmic and 12\% on Splice. In \textit{farfirst}, the K-LID-SVM outperforms the other defenses in all scenarios except Ijcnn1 and OMNeT, where LS-SVM has $7.4\%$ and $5.9\% $lower average error rates respectively. In \textit{nearest}, LS-SVM outperforms K-LID-SVM on Seismic by $0.85\%$ and LN-SVM outperforms K-LID-SVM on Acoustic by $0.7\%$. In all the other datasets we observe that K-LID-SVM outperforms the other defenses by large margins (up to 15\% against \textit{farfirst} and up to 25\% against \textit{nearest}). Figure \ref{fig:naive_error_rates} shows how the error rates vary on MNIST, Acoustic and OMNeT when the training flip rate increases from 0\% to 30\%.

\begin{figure*}[]
	\centering
	\begin{subfigure}{0.3\textwidth} 
		\includegraphics[width=\textwidth]{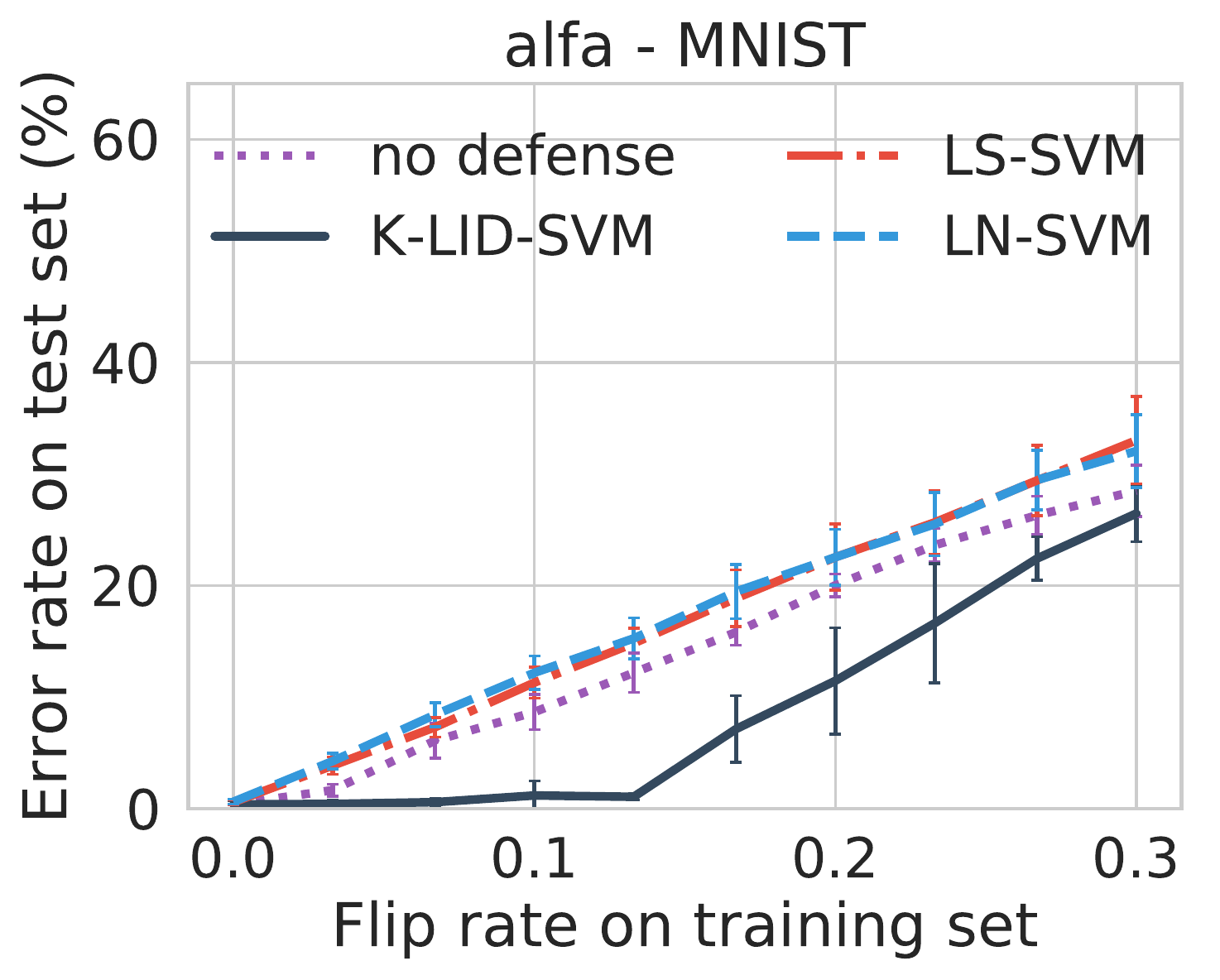}
		\label{fig:mnist_alfa}
	\end{subfigure}
	\begin{subfigure}{0.3\textwidth} 
		\includegraphics[width=\textwidth]{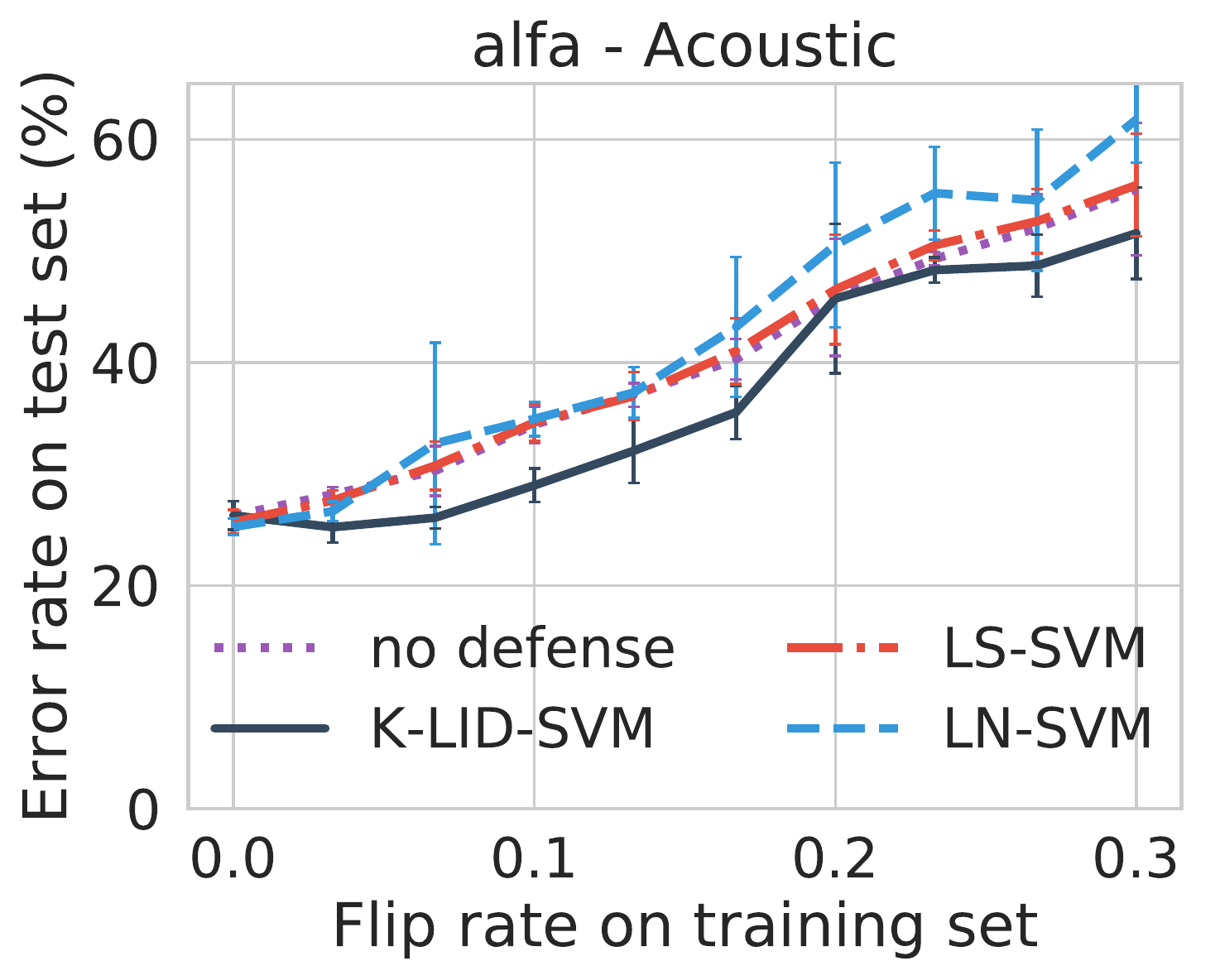}
		\label{fig:acoustic_alfa}
	\end{subfigure}
	\begin{subfigure}{0.3\textwidth} 
		\includegraphics[width=\textwidth]{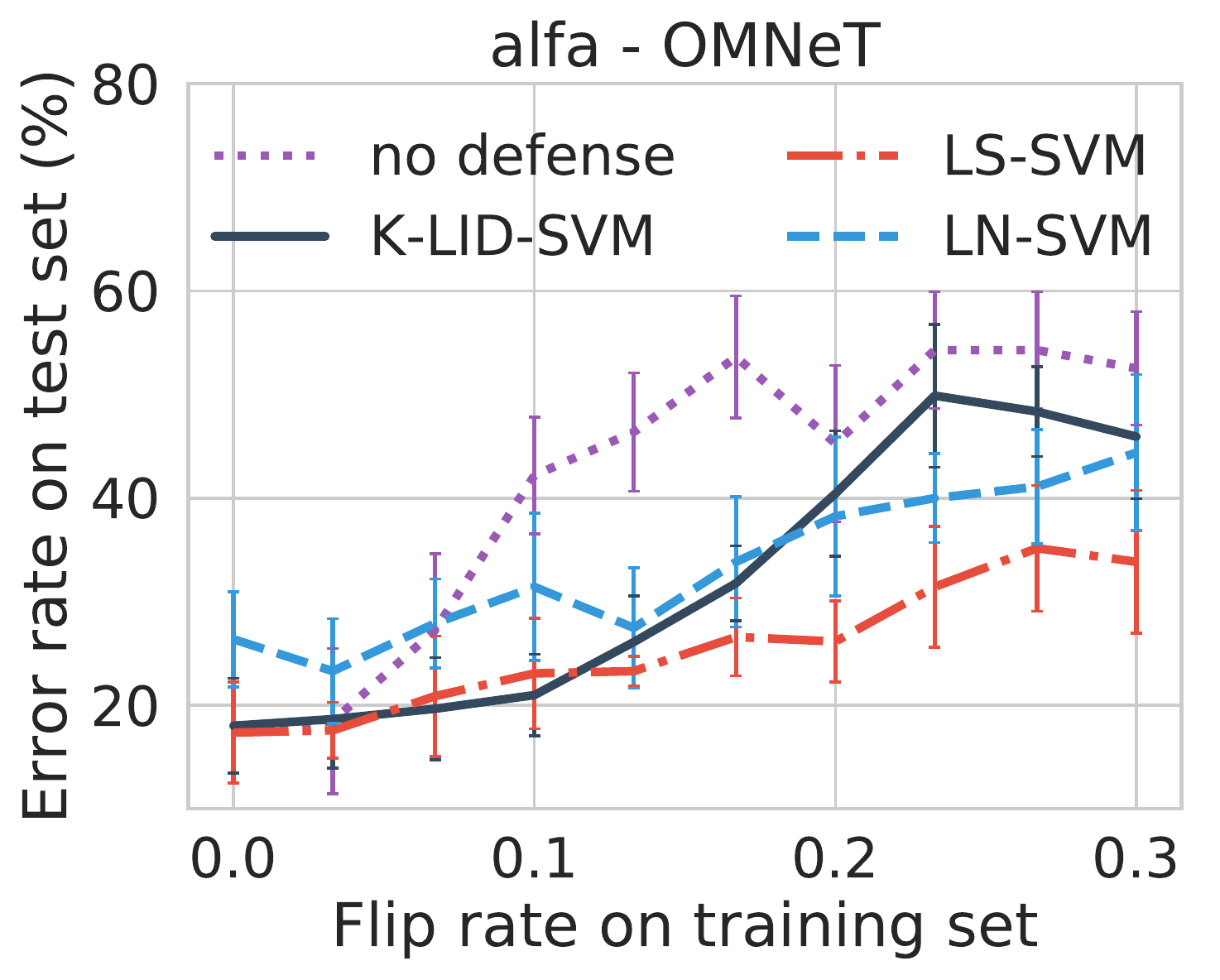}
		\label{fig:omnet_alfa}
	\end{subfigure}
	\begin{subfigure}{0.3\textwidth} 
		\includegraphics[width=\textwidth]{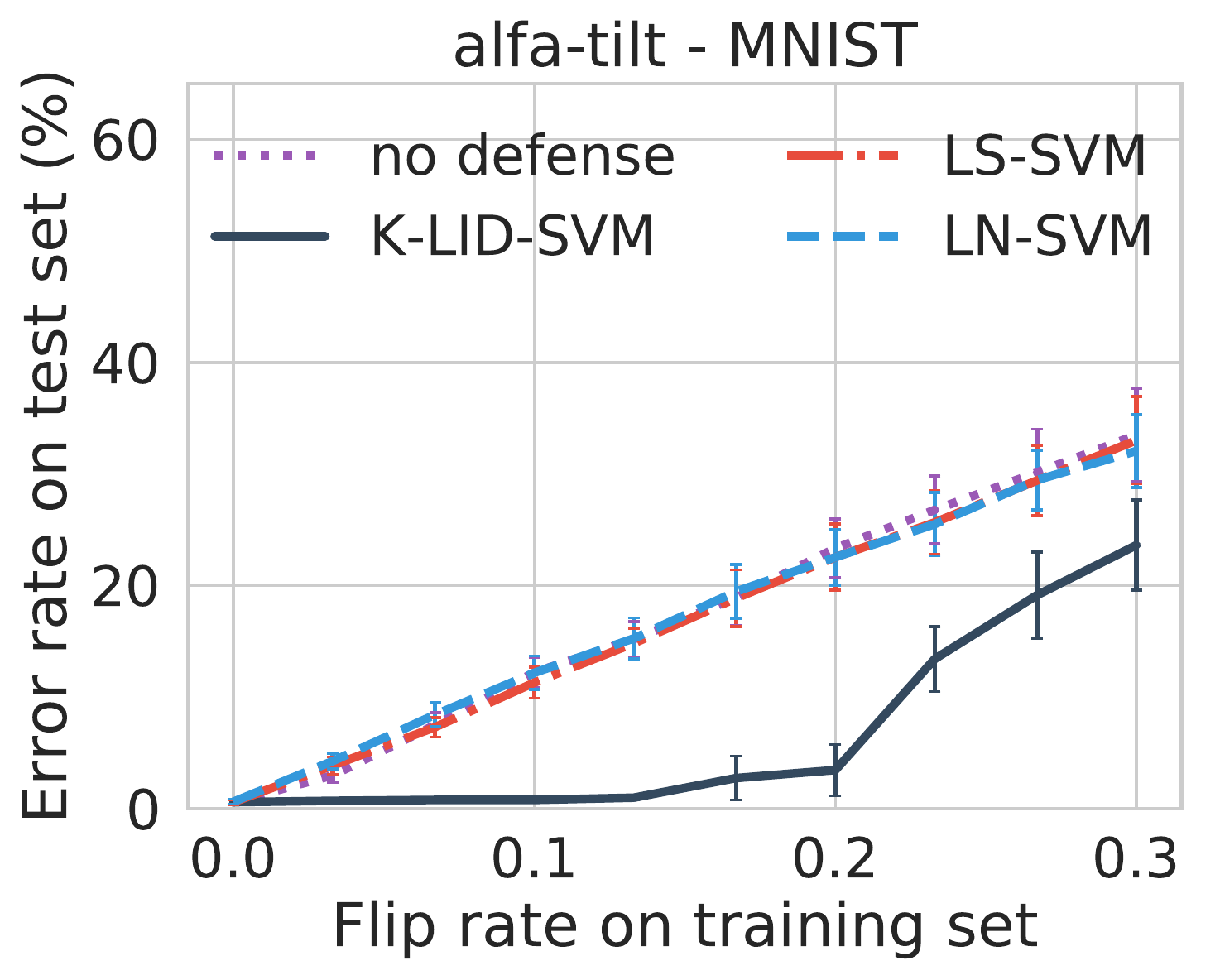}
		\label{fig:mnist_alfa-tilt}
	\end{subfigure}
	\begin{subfigure}{0.3\textwidth} 
		\includegraphics[width=\textwidth]{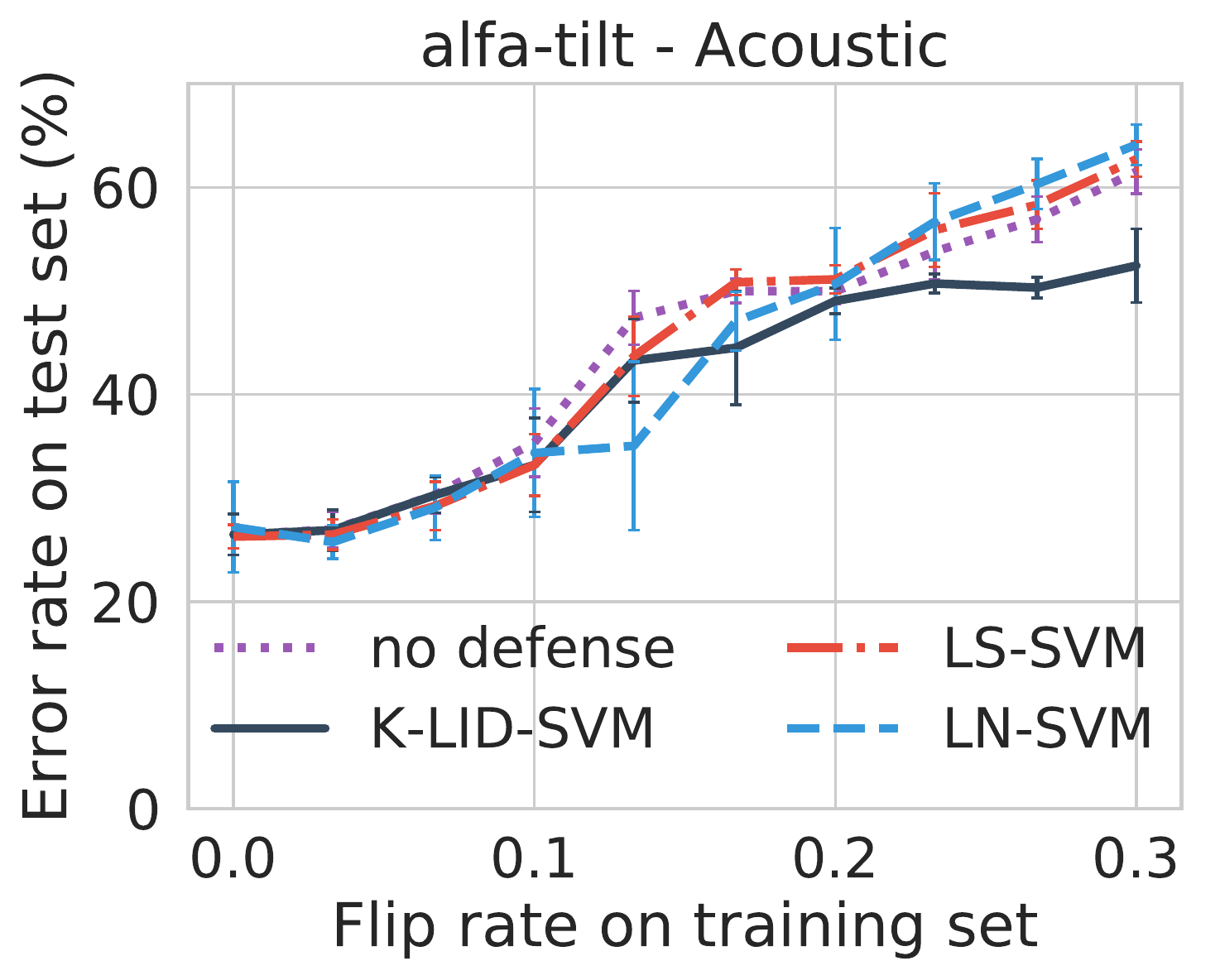}
		\label{fig:acoustic_alfa-tilt}
	\end{subfigure}
	\begin{subfigure}{0.3\textwidth} 
		\includegraphics[width=\textwidth]{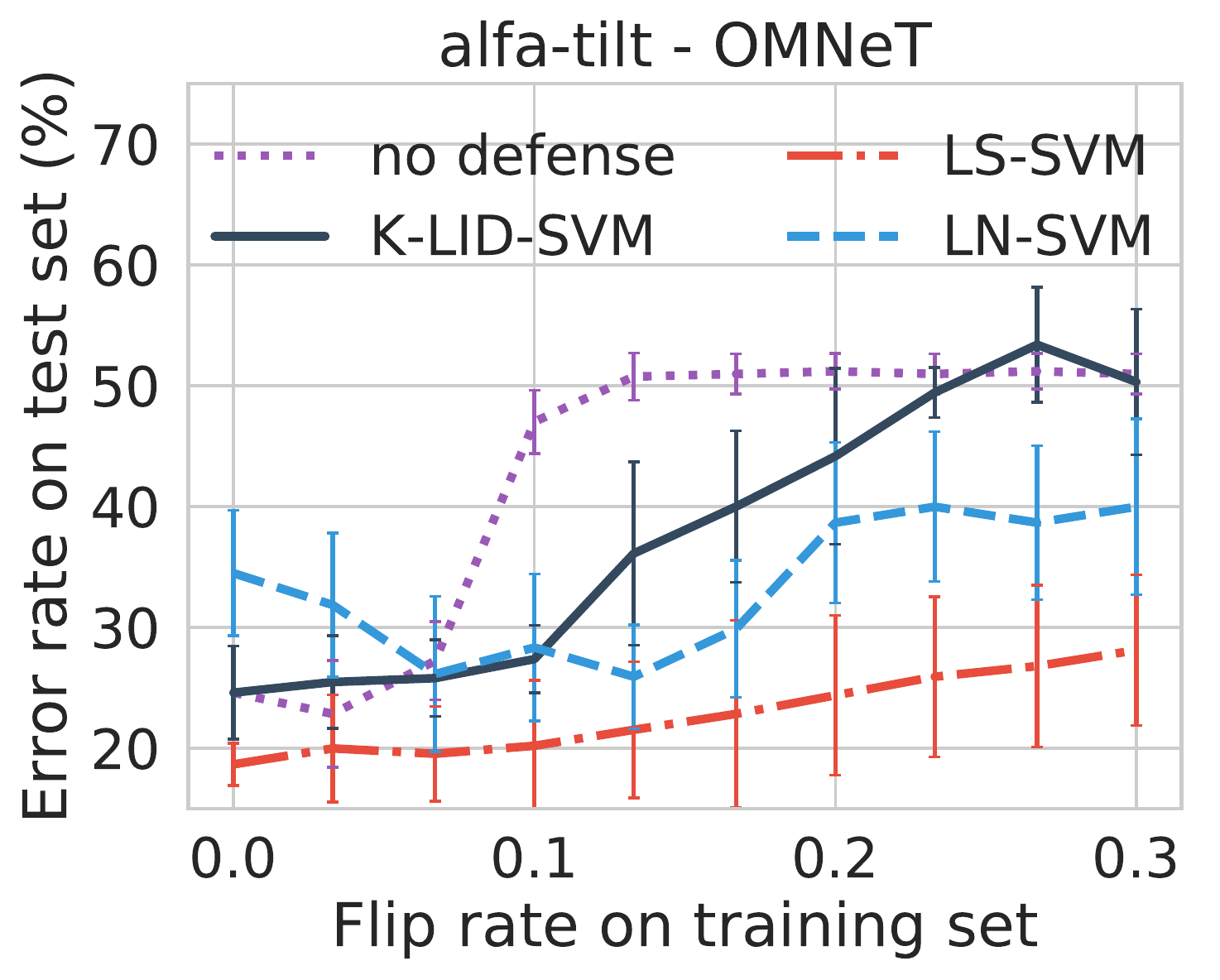}
		\label{fig:omnet_alfa-tilt}
	\end{subfigure}
	\caption{The average error rates of SVM, K-LID-SVM, LS-SVM and LN-SVM ($\mu=0.15$) against sophisticated label flip attacks on MNIST, Acoustic and OMNeT when the training flip rate increases from 0\% to 30\%.}
	\label{fig:sophisticated_error_rates}
\end{figure*}
\textbf{Sophisticated adversarial label flips:} Carefully selected adversarial label flips have a significant impact on the performance of SVMs. Against \textit{alfa}, the error rates go up by 28\% on MNIST, 34\% on OMNeT, 29\% on Acoustic, 40\% on Ijcnn1, 24\% on Seismic and 28\% on Splice when the flip rate is increased from 0\% to 30\%. Against \textit{alfa-tilt}, the error rate increases are 33\% on MNIST, 35\% on OMNeT, 35\% on Acoustic, 39\% on Ijcnn1, 33\% on Seismic and 27\% on Splice. Under alfa and alfa-tilt attacks, LS-SVM outperforms K-LID-SVM on the Ijcnn1 and dataset with $5.05\%$ and $1.05\%$ lower error rates respectively. Similarly, LS-SVM has $1\%$ and $15\%$ lower error rates on the OMNeT dataset under the same attacks. On all other datasets, the K-LID-SVM can significantly reduce the error rates compared to the other two defenses against all attack strategies. We observe that, on average, K-LID-SVM can achieve $1.3\%$ to $10.5\%$ lower error rates compared to the undefended SVM, $1.4\%$ to $16.7\%$ lower error rates compared to LN-SVM and up to $10.1\%$ lower error rates compared to LS-SVM. Figure \ref{fig:sophisticated_error_rates} shows the impact of \textit{alfa} and \textit{alfa-tilt} on three datasets.

\begin{table}[]
	\singlespacing
	\centering
	\caption{Average error rates across all the attack rates of each defense algorithm against the three attacks considered. The best results are indicated in \textbf{bold} font.}
	\label{tab:avg_error_rates_perturb}
	\begin{tabular}{@{}clr>{\columncolor[gray]{0.9}}rrr@{}}
		\toprule
		& Dataset  & SVM & K-LID-SVM & LS-SVM & CURIE \\\midrule
		{\parbox[t]{2mm}{\multirow{5}{*}{\rotatebox[origin=l]{90}{PA}}}}    & MNIST    & 0.68  & 0.65  & \textbf{0.60}  & 0.64  \\
		& Acoustic & 29.61 & \textbf{28.12} & 28.78 & 28.24 \\
		& Ijcnn1   & 16.42 & 15.51 & \textbf{12.07} & 16.36 \\
		& Seismic  & 22.52 & \textbf{21.23} & 22.11 & 22.33 \\
		& Splice   & 20.94 & \textbf{18.65} & 19.16 & 21.53 \\
		& OMNeT    & 30.38 & 29.47   		& \textbf{28.29} & 30.44 \\\midrule
		{\parbox[t]{2mm}{\multirow{5}{*}{\rotatebox[origin=c]{90}{RA}}}}  & MNIST    & 0.84  & 0.78  & \textbf{0.57}  & 0.73  \\
		& Acoustic & 31.74 & 29.13 & 31.16 & \textbf{28.66} \\
		& Ijcnn1   & 15.63 & 12.86 & \textbf{9.43}  & 13.80 \\
		& Seismic  & 24.77 & \textbf{18.24} & 18.61 & 20.14 \\
		& Splice   & 20.39 & \textbf{17.16} & 18.80 & 20.91 \\
		& OMNeT    & \textbf{18.20} & 18.89 & 19.32 & 18.86 \\\midrule
		{\parbox[t]{2mm}{\multirow{5}{*}{\rotatebox[origin=c]{90}{CG}}}}   & MNIST    & 0.68  & 0.66  & \textbf{0.56}  & 0.68  \\
		& Acoustic & 31.26 & \textbf{28.42} & 30.68 & 30.41 \\
		& Ijcnn1   & 16.07 & \textbf{13.66} & 15.72 & 14.99 \\
		& Seismic  & 23.43 & \textbf{18.08} & 19.24 & 20.49 \\
		& Splice   & 19.76 & \textbf{16.09} & 17.72 & 18.88 \\
		& OMNeT    & 18.02 & 18.70   		& \textbf{17.36} & 18.90 \\
		\bottomrule
	\end{tabular}
\end{table}

\begin{figure*}[]
	\centering
	\begin{subfigure}{0.3\textwidth} 
		\includegraphics[width=\textwidth]{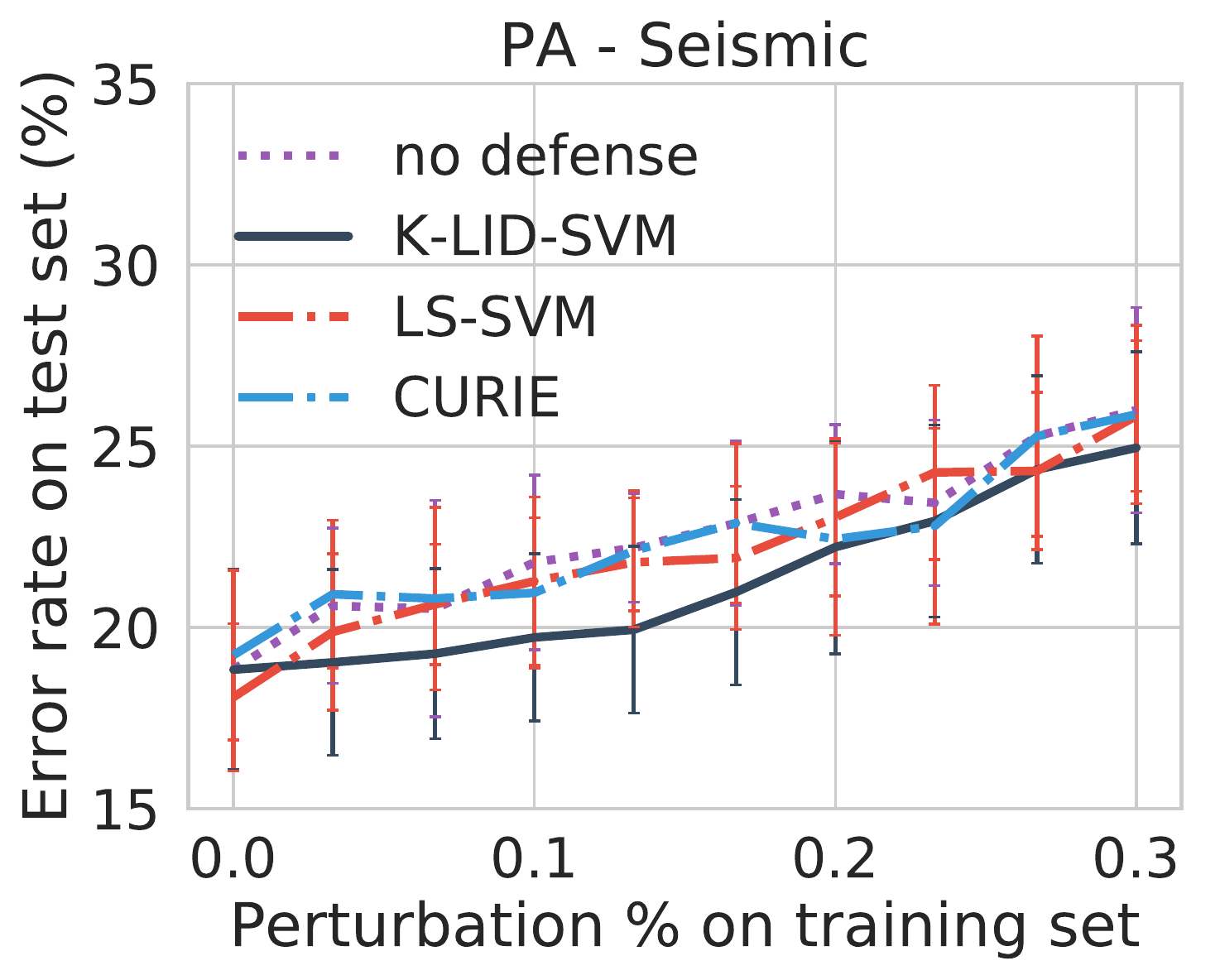}
		\label{fig:seismic_poison}
	\end{subfigure}
	\begin{subfigure}{0.3\textwidth} 
		\includegraphics[width=\textwidth]{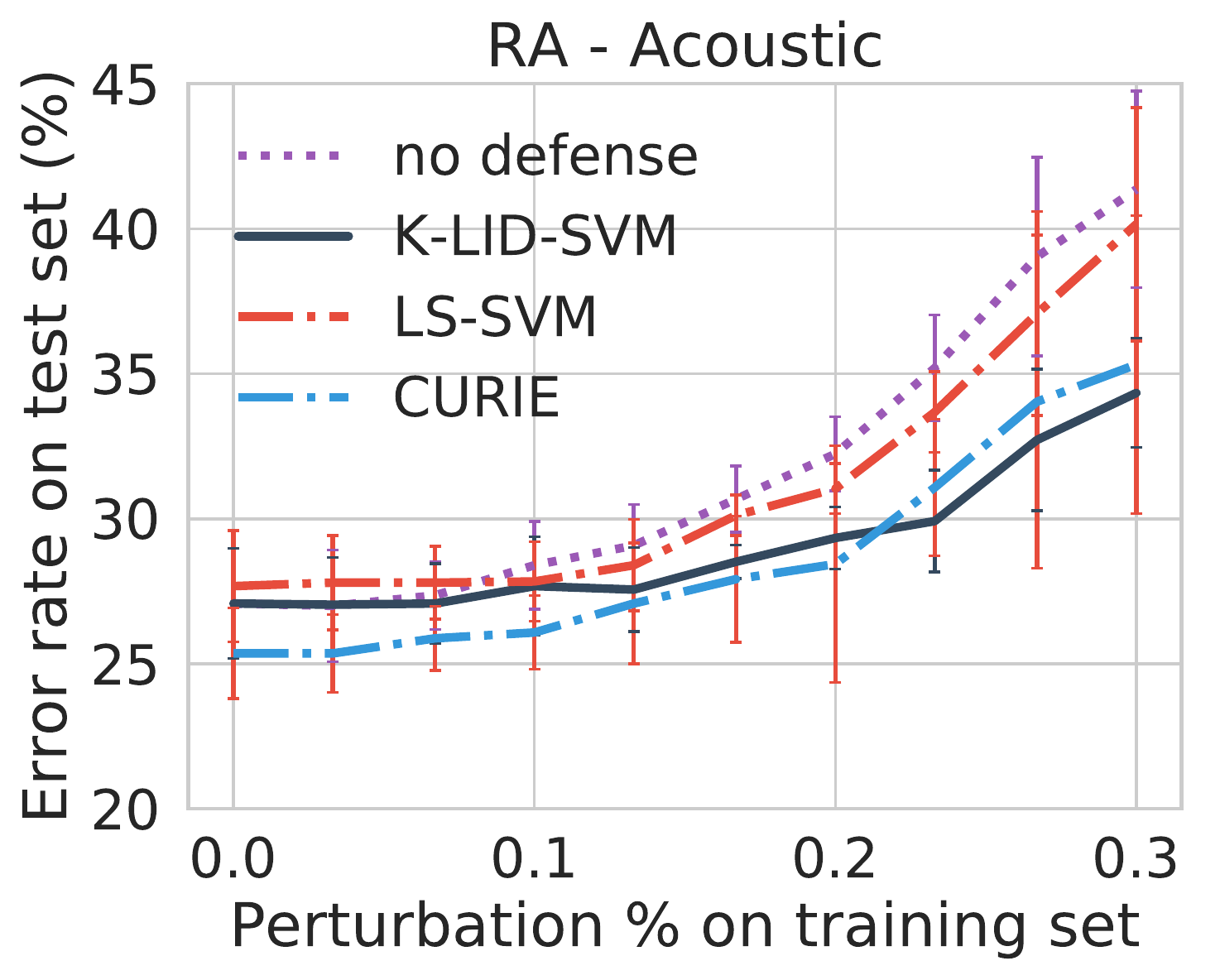}
		\label{fig:acoustic_ra}
	\end{subfigure}
	\begin{subfigure}{0.3\textwidth} 
		\includegraphics[width=\textwidth]{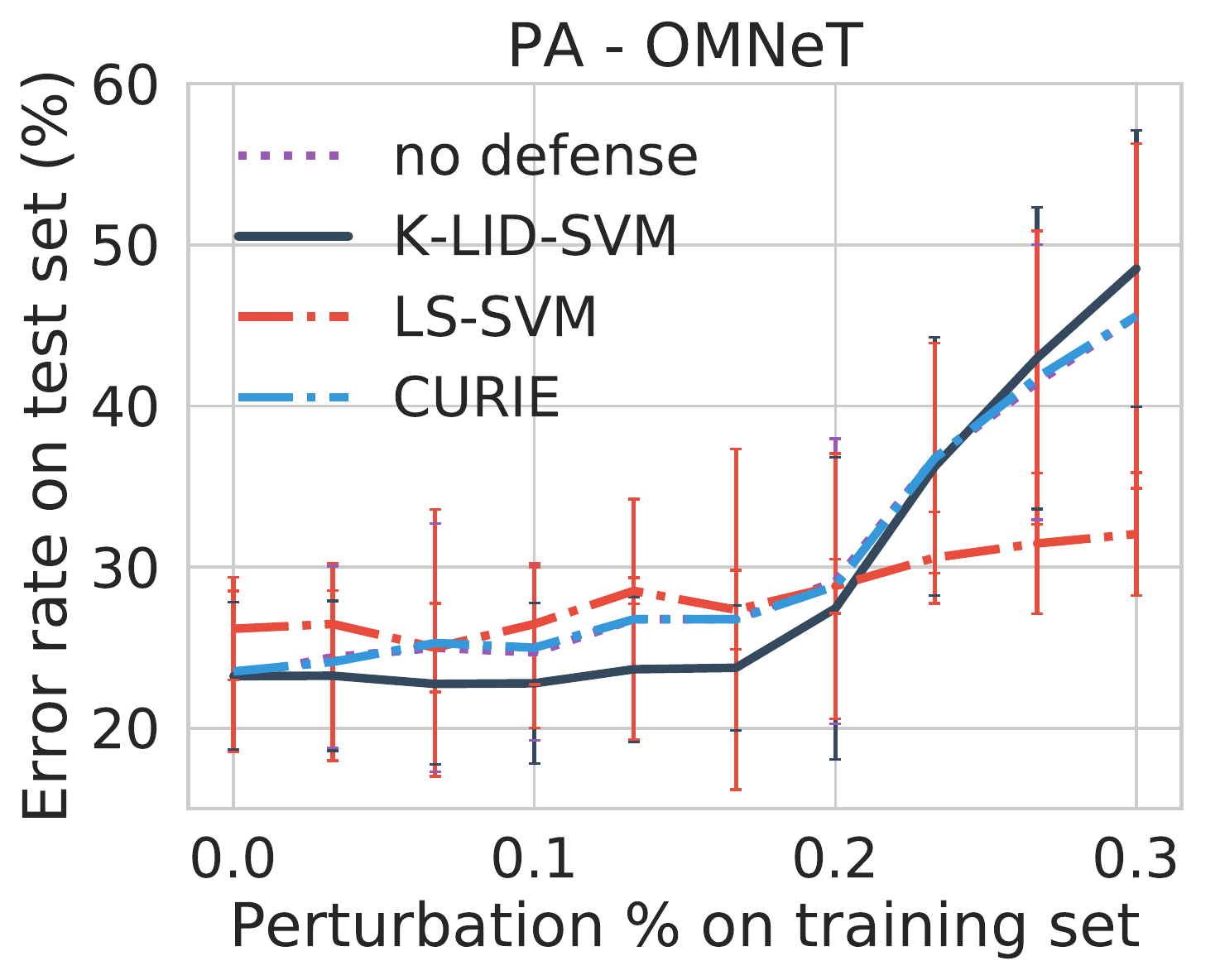}
		\label{fig:omnet_poison}
	\end{subfigure}
	\caption{The average error rates of SVM, K-LID-SVM and LS-SVM against poisoning attacks on three datasets when the attack percentage increases from 0\% to 30\%.}%
	\label{fig:all_error_rates_poison}
\end{figure*}

\subsubsection{Under Poisoning Attacks}
We compare the performance of K-LID-SVM against LS-SVM \cite{suykens1999least} and CURIE \cite{laishram2016curie} under poisoning attacks. The two algorithms use different approaches to address the problem of learning under adversarial conditions. Table \ref{tab:avg_error_rates_perturb} gives the error rate of each defense mechanism averaged over all the poison rates considered ($0\%$ - $30\%$).

Similar to the \textit{random} label flip attack, the performance of the binary SVM without a defense against poisoning attacks varies from dataset to dataset. On MNIST, we observe that it can retain a near $1\%$ error rate under all three attack algorithms considered. On the other data sets, however, we see a considerable impact on detection accuracy. 

Under \textit{PA}, the error rates increase by $22.3\%$ on OMNeT, $3.7\%$ on Acoustic, $10.5\%$ on Ijcnn1, $7.2\%$ on Seismic and $4.7\%$ on Splice when the perturbation rate is increased from $0\%$ to $30\%$. The LS-SVM outperforms K-LID-SVM on Ijcnn1 and OMNeT with $3.44\%$ and $1.2\%$ lower average error rates respectively. On the other data sets, however, K-LID-SVM outperforms the other defense algorithms by up to $2.9\%$.

Under \textit{RA}, we see a $0.2\%$ increase in error rate on MNIST, $0.9\%$ on OMNeT, $14.3\%$ on Acoustic, $10.3\%$ on Ijcnn1, $5.4\%$ on Seismic and $5.8\%$ on Splice when the perturbation rate goes from $0\%$ to $30\%$. Again we see LS-SVM outperforming K-LID-SVM on Ijcnn1 with a $3.4\%$ lower average error rate. We also observe CURIE having a $0.47\%$ lower average error rate on Acoustic. On the Seismic and Splice datasets, however, K-LID-SVM outperforms the other defenses by up to $3.8\%$ lower error rates. 

We observe that under \textit{CG}, the K-LID-SVM can consistently outperform LS-SVM and CURIE with lower average error rates up to $2.8\%$ on all the considered datasets except for OMNeT where LS-SVM has a $1.3\%$ lower average error rate. Figure \ref{fig:all_error_rates_poison} shows the impact of \textit{PA} and \textit{RA} on Seismic, Acoustic and OMNeT when the training perturbation rate increases from $0\%$ to $30\%$.

\subsection{Discussion}
We observe that the adversarial attacks such as \textit{farfirst}, \textit{alfa} and \textit{alfa-tilt} increase the error rates of the tested learners considerably compared to random label flips. This shows that although SVMs may be able to handle label noise in some scenarios by design, they are not immune to adversarial label flip attacks and by selectively flipping labels, adversaries can significantly increase the error rates. The \textit{nearest} attack, where the labels of data points that are nearest to the separating hyperplane are flipped, has the least impact on the prediction accuracy of learners across all the considered test cases. We speculate that flipped labels near the hyperplane results in less movement/rotation in the margin compared to label flips that are farther away.

From the three poisoning attacks we have considered in this paper, \textit{PA} \cite{Biggio:2012:PAA:3042573.3042761} and \textit{CG} \cite{li2016general} are unable to perform simultaneous optimization of multi-point attacks (i.e., collectively perturb data points such that there is a maximal increase in validation error). Furthermore, \textit{PA} \cite{Biggio:2012:PAA:3042573.3042761} attempts to find a reasonably good local maximum of the non-convex validation error surface, which may result in sub optimal attacks. Therefore we see in Table \ref{tab:avg_error_rates_perturb} that the impact of poisoning attacks on SVMs is similar to that of \textit{random} and \textit{nearest} label flip attacks.

Although LS-SVM, LN-SVM and CURIE add some resistance to SVMs against training time attacks, K-LID-SVM can consistently reduce error rates across different attack strategies and datasets. LN-SVM and LS-SVM try to address the problem of learning under adversarial attacks by spreading the influence on the decision boundary more evenly across all samples (using heuristics), whereas in K-LID-SVM we make samples that are suspected to be attacked contribute less. From the extensive experiments conducted, we see that the novel K-LID calculation we introduce has the potential to distinguish attacked samples from benign samples and thereby subdue the malicious actions of adversaries. 

CURIE attempts to learn under adversarial conditions by filtering out data points injected by the adversary using an algorithm based on clustering. The authors claim that attacked samples stand out from non-attack samples in (feature + label) space. As filtering is a pre-processing step that happens before training, CURIE can be used in conjunction with K-LID-SVM to further improve the attack resistance. 

A distributed SVM based learning solution allows for lower communication overhead without significantly compromising detection accuracy as shown in Section \ref{sec:dist_vs_cent}. Therefore it is ideal to be used in the SDR based cognitive radio network to detect malicious transmission sources. But the main drawback of the distributed learning system is that it exposes multiple entry points for attackers and an attack can propagate through the network even if a single node is compromised. As demonstrated by the experimental results, using K-LID-SVM would facilitate secure distributed detection using SDRs while benefiting from the reduced communication overhead provided by the DSVM framework.

Although no significant differences in terms of running times were observed during the above experiments, further research could be conducted to determine the relative efficiency of the different defense algorithms. While mini-batch sampling is a tested method for improving the efficiency of K-LID-SVM \cite{amsaleg2015estimating, LID_sarah_ICLR}, there is room for significant improvement through parallelization.

In summary, the experiments demonstrate that (i) SVMs are vulnerable to adversarial label flip attacks and poisoning attacks, (ii) LID values in the input space may not have sufficient distinguishing power when the data from the two classes are not linearly separable (whereas K-LID does), (iii) K-LID-SVM can withstand label flipping attacks as well as poisoning attacks (iv) de-emphasizing the effect of suspected samples gives better performance than methods that attempt to make all samples contribute to the decision process (e.g., LS-SVM and LN-SVM), (v) distributed detection using a DSVM framework has less communication overhead compared to a centralized learner under adversarial conditions.

\section{Conclusions}\label{sec:conclusions}
In this paper, we have addressed the challenge of increasing the attack resistance of SVMs against adversarial training time attacks. We observed that carefully crafted label flips and perturbations can significantly degrade the classification performance of SVMs. We introduced a novel LID approximation (K-LID) that makes use of the kernel matrix to obtain the LID values as well as three different label dependent variations of K-LID that can be used in situations with label flips. Using the K-LID, we proposed a weighted SVM (K-LID-SVM) and showed by testing against different attacks on several real-world datasets that it can be successfully utilized against label flip attacks as well as poisoning attacks. While there were some instances where LS-SVM, LN-SVM and CURIE outperformed K-LID-SVM, we observed that K-LID-SVM can achieve a higher level of stability across the different attacks and datasets considered in this evaluation. We observed that by using K-LID-SVM in a distributed manner, the learner can significantly reduce the communication overhead without sacrificing the classification accuracy. A further study could assess the feasibility of integrating K-LID into attack algorithms and evaluating the trade-off between the attack's severity and detectability.


\ifCLASSOPTIONcaptionsoff
  \newpage
\fi




\bibliographystyle{IEEEtran}
\bibliography{references}

\begin{thebibliography}{10}
\providecommand{\url}[1]{#1}
\csname url@samestyle\endcsname
\providecommand{\newblock}{\relax}
\providecommand{\bibinfo}[2]{#2}
\providecommand{\BIBentrySTDinterwordspacing}{\spaceskip=0pt\relax}
\providecommand{\BIBentryALTinterwordstretchfactor}{4}
\providecommand{\BIBentryALTinterwordspacing}{\spaceskip=\fontdimen2\font plus
\BIBentryALTinterwordstretchfactor\fontdimen3\font minus
  \fontdimen4\font\relax}
\providecommand{\BIBforeignlanguage}[2]{{%
\expandafter\ifx\csname l@#1\endcsname\relax
\typeout{** WARNING: IEEEtran.bst: No hyphenation pattern has been}%
\typeout{** loaded for the language `#1'. Using the pattern for}%
\typeout{** the default language instead.}%
\else
\language=\csname l@#1\endcsname
\fi
#2}}
\providecommand{\BIBdecl}{\relax}
\BIBdecl

\bibitem{vorobeychik2018adversarial}
Y.~Vorobeychik and M.~Kantarcioglu, ``Adversarial machine learning,''
  \emph{Synthesis Lectures on Artificial Intelligence and Machine Learning},
  pp. 1--169, 2018.

\bibitem{dalvi2004adversarial}
N.~Dalvi, P.~Domingos, S.~Sanghai, D.~Verma \emph{et~al.}, ``{Adversarial
  classification},'' in \emph{{10th ACM SIGKDD International Conference on
  Knowledge Discovery and Data Mining}}, 2004, pp. 99--108.

\bibitem{Biggio:2012:PAA:3042573.3042761}
B.~Biggio, B.~Nelson, and P.~Laskov, ``{Poisoning Attacks Against Support
  Vector Machines},'' in \emph{{29th International Coference on Machine
  Learning}}, ser. {ICML'12}, 2012, p. 1467–1474.

\bibitem{esmaeilpour2019robust}
M.~Esmaeilpour, P.~Cardinal, and A.~L. Koerich, ``A robust approach for
  securing audio classification against adversarial attacks,'' \emph{IEEE
  Transactions on Information Forensics and Security}, 2019.

\bibitem{sun2002road}
Z.~Sun, G.~Bebis, and R.~Miller, ``On-road vehicle detection using gabor
  filters and support vector machines,'' in \emph{2002 14th International
  Conference on Digital Signal Processing Proceedings. DSP 2002 (Cat. No.
  02TH8628)}, vol.~2.\hskip 1em plus 0.5em minus 0.4em\relax IEEE, 2002, pp.
  1019--1022.

\bibitem{steinhardt2017certified}
J.~Steinhardt, P.~W.~W. Koh, and P.~S. Liang, ``{Certified defenses for data
  poisoning attacks},'' in \emph{{Advances In Neural Information Processing
  Systems}}, 2017, p. 3517–3529.

\bibitem{suykens1999least}
J.~A. Suykens and J.~Vandewalle, ``{Least squares support vector machine
  classifiers},'' \emph{Neural Processing Letters}, vol.~9, no.~3, pp.
  293--300, 1999.

\bibitem{biggio2011support}
B.~Biggio, B.~Nelson, and P.~Laskov, ``{Support vector machines under
  adversarial label noise},'' in \emph{{Asian Conference on Machine Learning}},
  2011, pp. 97--112.

\bibitem{amsaleg2017vulnerability}
L.~Amsaleg, J.~Bailey, D.~Barbe, S.~Erfani, M.~E. Houle, V.~Nguyen, and
  M.~Radovanovi{\'c}, ``The vulnerability of learning to adversarial
  perturbation increases with intrinsic dimensionality,'' in \emph{2017 IEEE
  Workshop on Information Forensics and Security (WIFS)}, 2017, pp. 1--6.

\bibitem{LID_sarah_ICLR}
X.~Ma, B.~Li, Y.~Wang, S.~M. Erfani, S.~N.~R. Wijewickrema, G.~Schoenebeck,
  D.~Song, M.~E. Houle, and J.~Bailey, ``Characterizing adversarial subspaces
  using local intrinsic dimensionality,'' in \emph{6th International Conference
  on Learning Representations, {ICLR} 2018}, 2018.

\bibitem{LID1_Houle}
M.~E. Houle, ``{Local Intrinsic Dimensionality I: An Extreme-Value-Theoretic
  Foundation for Similarity Applications},'' in \emph{{Similarity Search and
  Applications}}, C.~Beecks, F.~Borutta, P.~Kr{\"o}ger, and T.~Seidl, Eds.,
  2017, pp. 64--79.

\bibitem{LID2_Houle}
{Houle, Michael E}, ``{Local intrinsic dimensionality II: multivariate analysis
  and distributional support},'' in \emph{{International Conference on
  Similarity Search and Applications}}, 2017, pp. 80--95.

\bibitem{pmlr-v80-ma18d}
X.~Ma, Y.~Wang, M.~E. Houle, S.~Zhou, S.~Erfani, S.~Xia, S.~Wijewickrema, and
  J.~Bailey, ``{Dimensionality-Driven Learning with Noisy Labels},'' in
  \emph{{35th International Conference on Machine Learning}}, ser. {Proceedings
  of Machine Learning Research}, vol.~80, 2018, pp. 3355--3364.

\bibitem{zhang2017}
{Zhang, Rui and Zhu, Quanyan}, ``{A game-theoretic analysis of label flipping
  attacks on distributed support vector machines},'' in \emph{{2017 51st Annual
  Conference on Information Sciences and Systems (CISS)}}.\hskip 1em plus 0.5em
  minus 0.4em\relax IEEE, 2017, pp. 1--6.

\bibitem{wang2012distributed}
D.~Wang and Y.~Zhou, ``Distributed support vector machines: An overview,'' in
  \emph{2012 24th Chinese Control and Decision Conference (CCDC)}.\hskip 1em
  plus 0.5em minus 0.4em\relax IEEE, 2012, pp. 3897--3901.

\bibitem{alpcan2009distributed}
T.~Alpcan and C.~Bauckhage, ``{A distributed machine learning framework},'' in
  \emph{{48th IEEE Conference on Decision and Control, 2009 held jointly with
  the 2009 28th Chinese Control Conference. CDC/CCC 2009.}}\hskip 1em plus
  0.5em minus 0.4em\relax IEEE, 2009, pp. 2546--2551.

\bibitem{forero2010consensus}
P.~A. Forero, A.~Cano, and G.~B. Giannakis, ``Consensus-based distributed
  support vector machines,'' \emph{Journal of Machine Learning Research},
  vol.~11, no. May, pp. 1663--1707, 2010.

\bibitem{frenay2014classification}
B.~Fr{\'e}nay and M.~Verleysen, ``{Classification in the presence of label
  noise: a survey},'' \emph{IEEE Transactions on Neural Networks and Learning
  Systems}, vol.~25, no.~5, pp. 845--869, 2014.

\bibitem{biggio2014security}
B.~Biggio, G.~Fumera, and F.~Roli, ``Security evaluation of pattern classifiers
  under attack,'' \emph{IEEE Transactions On Knowledge And Data Engineering},
  vol.~26, no.~4, pp. 984--996, 2014.

\bibitem{nettleton2010study}
D.~F. Nettleton, A.~Orriols-Puig, and A.~Fornells, ``{A study of the effect of
  different types of noise on the precision of supervised learning
  techniques},'' \emph{Artificial Intelligence Review}, vol.~33, no.~4, pp.
  275--306, 2010.

\bibitem{libralon2009pre}
G.~L. Libralon, A.~C.~P. de~Leon~Ferreira, A.~C. Lorena \emph{et~al.},
  ``{Pre-processing for noise detection in gene expression classification
  data},'' \emph{Journal of the Brazilian Computer Society}, vol.~15, no.~1,
  pp. 3--11, 2009.

\bibitem{zhang2003robustness}
J.~Zhang and Y.~Yang, ``{Robustness of regularized linear classification
  methods in text categorization},'' in \emph{{26th Annual International ACM
  SIGIR Conference on Research and Development in Informaion Retrieval}}, 2003,
  pp. 190--197.

\bibitem{manwani2013noise}
N.~Manwani and P.~Sastry, ``{Noise tolerance under risk minimization},''
  \emph{IEEE Transactions on Cybernetics}, vol.~43, no.~3, pp. 1146--1151,
  2013.

\bibitem{an2013fuzzy}
W.~An and M.~Liang, ``{Fuzzy support vector machine based on within-class
  scatter for classification problems with outliers or noises},''
  \emph{Neurocomputing}, vol. 110, pp. 101--110, 2013.

\bibitem{natarajan2013learning}
N.~Natarajan, I.~S. Dhillon, P.~K. Ravikumar, and A.~Tewari, ``{Learning with
  noisy labels},'' in \emph{{Advances in Neural Information Processing
  Systems}}, 2013, pp. 1196--1204.

\bibitem{liu2016classification}
T.~Liu and D.~Tao, ``{Classification with noisy labels by importance
  reweighting},'' \emph{IEEE Transactions on Pattern Analysis and Machine
  Intelligence}, vol.~38, no.~3, pp. 447--461, 2016.

\bibitem{Zhou:2012:ASV:2339530.2339697}
Y.~Zhou, M.~Kantarcioglu, B.~Thuraisingham, and B.~Xi, ``{Adversarial Support
  Vector Machine Learning},'' in \emph{{18th ACM SIGKDD International
  Conference on Knowledge Discovery and Data Mining}}, 2012, pp. 1059--1067.

\bibitem{laishram2016curie}
R.~Laishram and V.~V. Phoha, ``Curie: A method for protecting {SVM} classifier
  from poisoning attack,'' \emph{arXiv preprint arXiv:1606.01584}, 2016.

\bibitem{alfa}
H.~Xiao, B.~Biggio, B.~Nelson, H.~Xiao, C.~Eckert, and F.~Roli, ``{Support
  vector machines under adversarial label contamination},''
  \emph{Neurocomputing}, vol. 160, pp. 53--62, 2015.

\bibitem{li2016general}
B.~Li, Y.~Vorobeychik, and X.~Chen, ``A general retraining framework for
  scalable adversarial classification,'' \emph{arXiv preprint
  arXiv:1604.02606}, 2016.

\bibitem{yang2007weighted}
X.~Yang, Q.~Song, and Y.~Wang, ``{A weighted support vector machine for data
  classification},'' \emph{International Journal of Pattern Recognition and
  Artificial Intelligence}, vol.~21, no.~05, pp. 961--976, 2007.

\bibitem{amsaleg2015estimating}
L.~Amsaleg, O.~Chelly, T.~Furon, S.~Girard, M.~E. Houle, K.-I. Kawarabayashi,
  and M.~Nett, ``{Estimating local intrinsic dimensionality},'' in \emph{{21th
  ACM SIGKDD International Conference on Knowledge Discovery and Data
  Mining}}.\hskip 1em plus 0.5em minus 0.4em\relax ACM, 2015, pp. 29--38.

\bibitem{houle2013dimensionality}
M.~E. Houle, ``Dimensionality, discriminability, density and distance
  distributions,'' in \emph{2013 IEEE 13th International Conference on Data
  Mining Workshops}.\hskip 1em plus 0.5em minus 0.4em\relax IEEE, 2013, pp.
  468--473.

\bibitem{silverman2018density}
B.~W. Silverman, \emph{{Density estimation for statistics and data
  analysis}}.\hskip 1em plus 0.5em minus 0.4em\relax Routledge, 2018.

\bibitem{rajasegarar2010centered}
S.~Rajasegarar, C.~Leckie, J.~C. Bezdek, and M.~Palaniswami, ``Centered
  hyperspherical and hyperellipsoidal one-class support vector machines for
  anomaly detection in sensor networks,'' \emph{IEEE Transactions on
  Information Forensics and Security}, pp. 518--533, 2010.

\bibitem{flowers2019evaluating}
B.~Flowers, R.~M. Buehrer, and W.~C. Headley, ``Evaluating adversarial evasion
  attacks in the context of wireless communications,'' \emph{IEEE Transactions
  on Information Forensics and Security}, pp. 1102--1113, 2019.

\bibitem{Varga:2008:OOS:1416222.1416290}
A.~Varga and R.~Hornig, ``{An Overview of the OMNeT++ Simulation
  Environment},'' in \emph{{1st International Conference on Simulation Tools
  and Techniques for Communications, Networks and Systems \& Workshops}}, 2008,
  p.~60.

\end{thebibliography}

\end{document}